\documentclass[sigconf,nonacm]{acmart}
\usepackage{subcaption}
\usepackage{booktabs}
\usepackage{multirow}
\usepackage{tabularx}
\usepackage{booktabs}
\usepackage{array}
\newcolumntype{P}[1]{>{\centering\arraybackslash}p{#1}}
\usepackage{pifont}
\newcommand{\fully}{\ding{52}}

\newcommand{\notso}{\ding{56}}
\usepackage{sidecap}
\usepackage{tikz}
\usetikzlibrary{calc, positioning} 
\usepackage{appendix}
\AtBeginDocument{%
  }
\usepackage{array}
\newcolumntype{P}[1]{>{\centering\arraybackslash}p{#1}}

\acmISBN{978-1-4503-XXXX-X/2018/06}

\usepackage{natbib}

\usepackage{algorithm}
\usepackage{algpseudocode}
\usepackage{array}
\newcolumntype{P}[1]{>{\centering\arraybackslash}p{#1}}
\usepackage{enumitem}
\setlist[itemize]{leftmargin=*, noitemsep}
\setlist[enumerate]{leftmargin=*, noitemsep}  
\begin{document}


\noindent

\title{OPUS-VFL: Incentivizing Optimal Privacy-Utility Tradeoffs in Vertical Federated Learning}





\author{Sindhuja Madabushi, Ahmad Faraz Khan, Haider Ali, Jin-Hee Cho}
\affiliation{%
  \institution{Virginia Polytechnic Institute and State University, USA}
  \country{}  
}
\email{{msindhuja, ahmadfk, haiderali, jicho}@vt.edu}

\begin{abstract}
Vertical Federated Learning (VFL) enables organizations with disjoint feature spaces but shared user bases to collaboratively train models without sharing raw data. However, existing VFL systems face critical limitations: they often lack effective incentive mechanisms, struggle to balance privacy-utility tradeoffs, and fail to accommodate clients with heterogeneous resource capabilities. These challenges hinder meaningful participation, degrade model performance, and limit practical deployment.  To address these issues, we propose \texttt{OPUS-VFL}, an \underline{O}ptimal \underline{P}rivacy-\underline{U}tility tradeoff \underline{S}trategy for VFL. \texttt{OPUS-VFL} introduces a novel, privacy-aware incentive mechanism that rewards clients based on a principled combination of model contribution, privacy preservation, and resource investment. It employs a lightweight leave-one-out (LOO) strategy to quantify feature importance per client, and integrates an adaptive differential privacy mechanism that enables clients to dynamically calibrate noise levels to optimize their individual utility.  Our framework is designed to be scalable, budget-balanced, and robust to inference and poisoning attacks. Extensive experiments on benchmark datasets (MNIST, CIFAR-10, and CIFAR-100) demonstrate that \texttt{OPUS-VFL} significantly outperforms state-of-the-art VFL baselines in both efficiency and robustness. It reduces label inference attack success rates by up to 20\%, increases feature inference reconstruction error (MSE) by over 30\%, and achieves up to 25\% higher incentives for clients that contribute meaningfully while respecting privacy and cost constraints. These results highlight the practicality and innovation of \texttt{OPUS-VFL} as a secure, fair, and performance-driven solution for real-world VFL.
\end{abstract}

\ccsdesc[300]{Security and privacy~Distributed systems security}
\ccsdesc[500]{Computing methodologies~Cooperation and coordination}
\ccsdesc[500]{Computing methodologies~Artificial intelligence}

\keywords{Vertical federated learning, incentive mechanism, privacy-utility tradeoff, adaptive differential privacy, client contribution evaluation}



\maketitle

\begingroup
\renewcommand\thefootnote{}
\footnotetext{\large\textbf{This work has been submitted to the IEEE for possible publication.
Copyright may be transferred without notice, after which this version may no longer be accessible.}}
\endgroup

\section{Introduction} \label{sec:intro}
Federated Learning (FL) has emerged as a promising paradigm to mitigate the privacy and security risks inherent in conventional centralized machine learning~\cite{zhang2021survey}.  Traditional approaches rely on aggregating data in a central location, exposing sensitive user information to significant threats. FL enables multiple organizations or clients to collaboratively train machine learning models without sharing their raw data~\cite{wen2023survey}. Instead, only model updates or relevant parameters are exchanged, allowing for improved model performance while preserving data locality and privacy~\cite{zhang2021survey, wen2023survey}.

Federated learning (FL) systems are commonly categorized into Horizontal FL (HFL) and Vertical FL (VFL), depending on how data is distributed across participating organizations. HFL assumes that all parties share the same feature space but possess different subsets of data samples. In contrast, VFL assumes that participants share a common sample space but hold disjoint subsets of features~\citep{liu2024vertical}. \textbf{Figure~\ref{fig:vfl-hfl-description}} illustrates the key difference in data partitioning schemes between these two FL settings.

\textbf{\bf Why VFL?} VFL is especially important in domains like healthcare and finance, where organizations (e.g., hospitals, banks) hold complementary information about the same individuals but cannot share raw data due to regulations such as HIPAA and GDPR. For instance, imaging centers may store diagnostic scans, while hospitals maintain electronic health records (EHRs)~\citep{liu2024vertical, yang2019federated}, and financial institutions may hold different features for shared customers~\citep{zhu2021pivodl}. VFL enables secure, collaborative learning under such constraints, where data heterogeneity and alignment are more complex than in HFL. Real-world VFL systems must also address practical challenges like incentivizing participation, managing resource disparities, and balancing privacy with model utility~\citep{cui2024survey, tan2023fraim, li2023fedsdg}.

\textbf{Despite its potential, (V)FL faces fundamental limitations that hinder real-world adoption.}  Equally important is the need for \textit{incentive mechanisms} to encourage participation from clients with high-quality data and limited resources. Without proper rewards, participants may be disincentivized, undermining the overall performance and sustainability of the FL ecosystem~\citep{cui2024survey}. While several incentive mechanisms have been developed for HFL~\citep{le2021incentive, Toyoda19, zhang2021incentive, tang-incentive-2021, ng2021hierarchical, yan2021fedcm, deng2022improving, Han22-tiff}, their applicability to the VFL setting remains largely unexplored.

\noindent\textbf{\texttt{OPUS-VFL} redefines incentive-compatible and privacy preservation learning for vertical federated systems.} We propose an \underline{O}ptimal \underline{P}rivacy-\underline{U}tility tradeoff \underline{S}trategy, \texttt{OPUS-VFL}, to address model performance, client privacy, and computational efficiency. At its core, \texttt{OPUS-VFL} uses a dynamic, privacy-aware incentive mechanism that rewards clients based on their contributions to accuracy, privacy, and resource usage. The server estimates each client’s marginal contribution using a lightweight leave-one-out (LOO) evaluation. Clients calibrate their privacy levels by injecting Gaussian noise through an adaptive differential privacy (DP) mechanism~\citep{abadi2016deep}, enabling personalized privacy-utility trade-offs. The server integrates these signals to compute reward allocations that promote sustained, meaningful participation.

\texttt{OPUS-VFL} ensures economic fairness through two key properties: \textit{budget balance}, maintaining a fixed total reward budget, and \textit{individual rationality}, guaranteeing rewards exceed each client’s training costs. These properties support sustainable collaboration. Overall, \texttt{OPUS-VFL} offers a practical, scalable, and resilient solution for vertical federated learning by unifying privacy, fairness, and performance within an incentive-compatible framework.

To recap \texttt{OPUS-VFL}'s technical merits, we summarize the following \textbf{key contributions} of this work:
\begin{itemize}
\item \textbf{Privacy-aware contribution evaluation:} \texttt{OPUS-VFL} introduces a dynamic, lightweight leave-one-out strategy to assess each client’s impact on model performance in every training round. It integrates a customizable differential privacy mechanism, enabling clients to control their privacy-utility tradeoff. Unlike prior methods requiring costly Shapley values~\cite{tan2023fraim, lu2022truthful} or retraining~\cite{khan2024using}, our approach achieves accurate contribution measurement with minimal overhead and improved privacy protection.
\begin{figure}[t]
\centering
\begin{subfigure}[b]{0.23\textwidth}
\centering
\includegraphics[width=\textwidth]{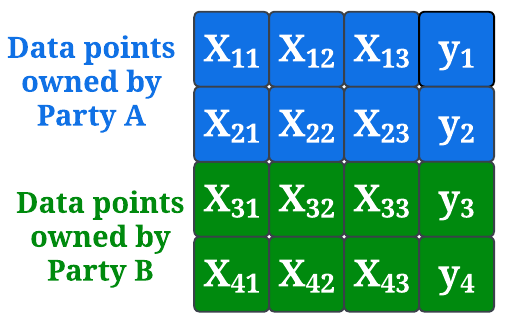}
\caption{Horizontal FL}
\end{subfigure}
\hfill
\begin{subfigure}[b]{0.23\textwidth}
\centering
\includegraphics[width=\textwidth]{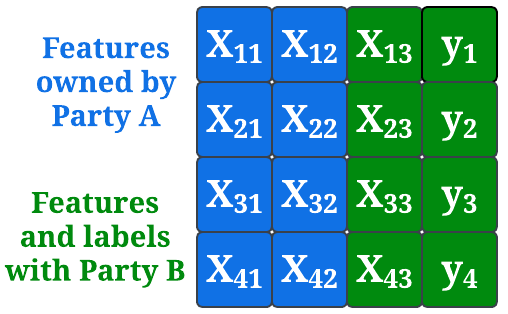}
\caption{Vertical FL}
\end{subfigure}
\hfill
\caption{\textbf{HFL vs. VFL:} HFL shares features across samples while VFL shares samples across features.}
\Description{OPUS-VFL motivation}
\label{fig:vfl-hfl-description}

\end{figure}

\item \textbf{Resource-sensitive incentive design:} Beyond performance contribution considered in prior works~\cite{deng-quality-aware-2021, Han22-tiff, ng2021multi, zhang-horizontal-2021}, \texttt{OPUS-VFL} incorporates \textit{contribution equity} by factoring in each client's resource investment during training. This joint consideration of data quality and resource cost ensures fairer compensation, encourages participation from clients with limited compute but valuable data, and better reflects real-world client heterogeneity.

\item \textbf{Scalability and efficiency:} \texttt{OPUS-VFL} avoids heavy
game theoretic~\cite{khan2024using, yang2023incentive} or Shapley-based~\cite{tan2023fraim, lu2022truthful} computations and remains efficient as the number of clients grows. A streamlined evaluation and aggregation protocol maintains robustness and feasibility in large-scale, heterogeneous VFL scenarios, particularly relevant in domains like healthcare and finance, where privacy constraints and client diversity are critical.
\end{itemize}


\begin{table*}[ht]
\centering
\caption{Comparison of VFL-based Systems With Respect to Incentive Mechanisms (IMs), Fairness, Privacy, and Key Techniques} \label{tab:vfl_incentive_comparison}
\begin{tabularx}{\textwidth}{@{}P{3.2cm}P{1.8cm}P{1.8cm}P{1.8cm}X@{}}
\toprule
\textbf{Scheme} & \textbf{IM} & \textbf{Fairness} & \textbf{Privacy} & \multicolumn{1}{c}{\textbf{Key Technique}} \\
\midrule
TEA~\cite{lu2022truthful} (2022) & \fully & \fully & \notso & Mechanism design, LP relaxation \\
FRAIM~\cite{tan2023fraim} (2023) & \fully & \fully & \fully & Shapley value, Hermite extrapolation \\
Khan et al.~\cite{khan2024using} (2024) & \fully & \fully & \notso & Nucleolus from cooperative game theory \\
Yang et al.~\cite{yang2023incentive} (2023) & \fully & \notso & \notso & Two-stage Stackelberg game \\
HiFi-Gas~\cite{sun2024hifi} (2024) & \fully & \fully & \notso & Hierarchical collaboration, reward design \\
ELXGB~\cite{xu2024elxgb} (2024) & \notso & \notso & \fully & DP in node splitting \\
PIVODL~\cite{zhu2021pivodl} (2021) & \notso & \notso & \fully & HE and DP for gradients and leaf values \\
FedSDG-FS++~\cite{li2023efficient} (2023) & \notso & \notso & \fully & Stochastic dual-gate, HE for label protection \\
VIM~\cite{xie2024improving} (2024) & \notso & \notso & \fully & Multi-head architecture, client-level DP \\
SecureVFL~\cite{fan2024securevfl} (2024) & \notso & \notso & \fully & Blockchain, 3-party secret sharing \\
FedVS~\cite{li2023fedvs} (2023) & \notso & \notso & \fully & Secret sharing \\
Falcon~\cite{wu2023falcon} (2023) & \notso & \notso & \fully & Threshold HE, secret sharing \\
Privet~\cite{zheng2023privet} (2023) & \notso & \notso & \fully & Lightweight cryptography \\
FedV~\cite{xu2021fedv} (2021) & \notso & \notso & \fully & Functional encryption \\
PraVFed~\cite{wang2025pravfed} (2025) & \notso & \notso & \fully & Blinding of local embeddings \\
OpenVFL~\cite{yang2024openvfl} (2024) & \notso & \notso & \fully & Private set intersection \\
Zhang et al.~\cite{zhang2024privacy} (2024) & \notso & \notso & \fully & Private set intersection \\
\textbf{OPUS-VFL (Ours)} & \fully & \fully & \fully & Leave-one-out, adaptive DP, tradeoff optimization \\
\bottomrule
\end{tabularx}

\fully: Considered; \notso: Not considered.
\end{table*}

\section{Related Work} \label{sec:related-work}
Although a greater number of incentive mechanisms (IMs) have been developed for horizontal federated learning (HFL), our focus lies in vertical federated learning (VFL). Therefore, this literature review primarily covers IMs designed for VFL systems, along with privacy-preserving techniques relevant to VFL settings.

\subsection{Incentive Mechanisms (IMs) in VFL}

\citet{lu2022truthful} proposed a method to ensure truthful data reporting in VFL by establishing a Nash equilibrium where truth-telling is the dominant strategy for all clients. The mechanism uses linear programming (LP) relaxations and sample-based computations to maintain incentive compatibility.  \citet{khan2024using} used cooperative game theory to allocate incentives based on the nucleolus, ensuring a fair distribution among clients. \citet{yang2023incentive} developed a model of the incentive interaction as a two-stage Stackelberg game, where a label owner optimizes reward parameters and data owners respond with their processing strategies to balance performance and reward maximization.  

A Shapley value-based method~\cite{tan2023fraim} was proposed to estimate client contributions based on feature importance in VFL. To improve scalability, Hermite extrapolation is used on sampled data, reducing full-data dependency while capturing marginal contributions effectively.  A hierarchical FL framework~\cite{sun2024hifi} integrates both horizontal and vertical FL settings, facilitating collaboration across organizational boundaries. It employs a multi-dimensional reward scheme that accounts for both data quality and model contribution to sustain engagement across diverse client types.

\textbf{Limitations and the Contributions of \texttt{OPUS-VFL}}  While the above methods introduce valuable mechanisms for client incentivization in VFL, several limitations remain. Many rely on complex game-theoretic or cooperative frameworks, incurring high computational costs and limiting scalability. Others lack empirical validation or practical adaptability in resource-constrained environments. Few methods effectively balance model utility with privacy preservation, nor do they dynamically adjust incentives based on both contribution quality and privacy sensitivity.

\texttt{OPUS-VFL} fills these gaps by proposing an efficient and scalable incentive mechanism that rewards clients based on a privacy-utility tradeoff. It leverages a leave-one-out (LOO) strategy to evaluate client contributions and applies adaptive differential privacy to preserve sensitive information while respecting resource constraints. By unifying fairness, performance, and privacy, OPUS-VFL ensures robust participation in VFL without incurring excessive computational or communication overhead.
\vspace{-0.15in}
\subsection{Privacy-Preserving VFL}

\subsubsection{\bf Cryptographic Privacy-Preserving VFL (PP-VFL)} Various cryptographic approaches have been explored to preserve privacy in VFL, as outlined below.

\textbf{Homomorphic Encryption (HE)-based Methods}
A class of encryption-based VFL approaches relies on HE to secure computation during model training. Methods such as~\cite{xu2024elxgb, zhu2021pivodl} integrate HE with differential privacy (DP) to protect sensitive information during node splitting and gradient boosting. Similarly, \citet{li2023efficient} extended FedSDG-FS++ with partial HE to secure label information at the server side.

\textbf{Secret Sharing and Blockchain-based Protocols}  Another line of work adopts secret sharing and blockchain to enhance data confidentiality and trust. For instance,~\citet{fan2024securevfl} employed a permissioned blockchain integrated with a three-party replicated secret sharing scheme to protect client data. Similarly, \citet{li2023fedvs} used secret sharing for privacy preservation, while~\citet{wu2023falcon} combined threshold partially HE with secret sharing to protect both training integrity and interpretability. Lightweight cryptographic solutions, such as~\cite{zheng2023privet}, aim to minimize overhead while ensuring gradient privacy in decision table models.

\textbf{Functional Encryption and Blinding-based Techniques}  Functional encryption and blinding methods have also been explored to reduce communication overhead and support secure gradient computation. For example,~\citet{xu2021fedv} utilized functional encryption to eliminate the need for peer-to-peer communication, expediting secure gradient updates.  \citet{wang2025pravfed} applied blinding factors to local embeddings, enhancing privacy while improving scalability in heterogeneous environments.

\textbf{Private Set Intersection Protocols}
Protocols based on private set intersection (PSI) align distributed datasets without revealing identity or label information. Approaches such as~\cite{yang2024openvfl, zhang2024privacy} incorporate PSI to protect client identities during data alignment and model training.

\textbf{Closing Gaps in Prior Work in PP-VFL: Contributions of OPUS-VFL}  While these cryptographic methods offer strong theoretical privacy guarantees, they often come with high computational and communication overhead, particularly those relying on HE or multi-party computation (MPC). Such overheads limit their practicality in real-world, resource-constrained VFL deployments. Additionally, these methods are typically designed to enforce privacy, rather than to integrate privacy with incentive alignment or performance-based contribution assessment.  \texttt{OPUS-VFL} addresses these limitations by adopting a lightweight, DP-based approach that enables privacy-preserving incentive mechanisms without expensive cryptographic operations. By combining \textit{leave-one-out contribution evaluation} with \textit{adaptive DP}, OPUS-VFL offers a scalable, efficient, and client-aware solution that jointly optimizes privacy protection and utility-based incentive allocation in vertical federated learning.

\subsubsection{\bf Differential Privacy (DP)-based VFL} 

\textbf{DP for Tree-based and Gradient Boosting Models}
\citet{xu2024elxgb} applied differential privacy in node splitting to protect sensitive information within XGBoost-based VFL. Similarly,~\citet{zhu2021pivodl} combined HE with DP to secure both gradients and leaf values in vertically partitioned gradient boosting models.

\textbf{DP in Embedding and Feature Perturbation}
\citet{li2023efficient} introduced the FedSDG-FS framework, which uses a Gaussian stochastic dual-gate mechanism to perturb local embeddings under DP constraints. Its extension, FedSDG-FS++, further incorporates partial homomorphic encryption to reinforce server-side protection of label information.

\textbf{Client-level DP in Model Architecture}
A client-level DP mechanism~\cite{xie2024improving} is adopted in a multi-head VFL architecture, allowing clients to perform multiple local updates before communication. This design enhances both privacy and model performance by reducing communication frequency and noise accumulation.

\textbf{Closing Gaps in Prior Work in DP: Contributions of OPUS-VFL}  While these approaches offer formal privacy guarantees by combining DP with cryptographic techniques, they often entail high computational and architectural complexity. Many depend on costly encryption, complex models, or centralized coordination, making them impractical for resource-constrained deployments. Moreover, they largely focus on privacy, neglecting incentive mechanisms tied to performance and resource usage. \texttt{OPUS-VFL} addresses these limitations using a lightweight, client-level differential privacy strategy that balances privacy preservation with performance-aware incentive allocation. Without relying on encryption, \texttt{OPUS-VFL} dynamically evaluates each client’s contribution via a leave-one-out (LOO) strategy and allocates rewards based on a principled utility-privacy tradeoff. This enables a scalable, practical, and incentive-compatible solution for privacy-preserving vertical federated learning.

\textbf{Table~\ref{tab:vfl_incentive_comparison}} summarizes representative VFL-based systems, comparing their support for incentive mechanisms, fairness, and privacy preservation, along with the key techniques employed. For our comparative performance analyses in Section~\ref{sec:results-analysis}, we include our proposed incentive schemes, Bid Price First (BPF)~\cite{tan2023fraim} and the Theoretically Optimal Organization Selection Mechanism (O2S)~\cite{tan2023fraim}, as well as privacy-preserving baselines: privacy-preserving deep learning (PPDL)~\cite{shokri2015privacy} and VFL-CZOFO~\cite{wang2023unified}. These methods are chosen for their lightweight designs, which support efficient computation compared to homomorphic encryption (HE)-based alternatives~\cite{li2023fedsdg, wu2023falcon, yang2024openvfl}, known for high computational overhead. We exclude TEA~\cite{lu2022truthful} due to the lack of open-source implementation and FRAIM~\cite{tan2023fraim} due to limited scalability and difficulties applying it to CIFAR-10. We also omit the scheme by \citet{yang2023incentive}, which primarily incentivizes processing speed—an irrelevant factor under our assumption of uniform capabilities. Similarly, HiFI-Gas~\cite{sun2024hifi}, and encryption-heavy methods like FedSDG-FS++~\cite{li2023fedsdg}, Falcon~\cite{wu2023falcon}, and OpenVFL~\cite{yang2024openvfl} are excluded due to significant runtime costs. Section~\ref{subsec:scalability-analysis} details our baseline selection rationale.

\section{Problem Statement} \label{sec:problem-statement}

The proposed \texttt{OPUS-VFL} establishes a scalable, equitable, and privacy-preserving incentive mechanism for VFL that balances client contributions, privacy constraints, and system cost. By dynamically evaluating feature importance and incorporating differential privacy (DP), \texttt{OPUS-VFL} ensures fair rewards while preserving client privacy. Its incentive mechanism promotes sustained participation by aligning incentives with meaningful contributions, optimizing the trade-off among performance, privacy, and cost-efficiency, and maintaining adaptability across diverse VFL architectures.

\texttt{OPUS-VFL} is formulated as a bi-level program, where the upper-level problem represents the global model and the lower-level problem represents the client models. Let $h = (h_1, h_2, \dots, h_N)$ be the activations of the $N$ clients, $\varepsilon = (\varepsilon_1, \varepsilon_2, \dots, \varepsilon_N)$ their privacy budgets (e.g., DP parameters), and $\mathcal{C} = (\mathcal{C}_1, \mathcal{C}_2, \dots, \mathcal{C}_N)$ their resource budgets (e.g., resource usage fractions). The global model solves the problem by:
\begin{align}
&\max_{\theta_g}  \: \mathrm{Acc}\big(\mathcal{M}(\theta_g, h)\big), \; \; 
&\:\:\: \text{s.t.} \: \sum_{i=1}^{N} \mathcal{R}_i(h, \varepsilon, \mathcal{C}, \theta_g) \leq \tau, \nonumber
\end{align}
where $\theta_g$ denotes the global model parameters, the objective denotes the global model's predictive performance (e.g., accuracy or negative loss), $\mathcal{R}_i$ is the nonnegative reward given by the server to client $i$, and $\tau$ is the server's reward budget.
The reward $\mathcal{R}_i$ weights the relative contribution of client $i$'s activation towards the global model accuracy with client $i$'s privacy and resource budgets.

The lower-level problem can be decomposed into $N$ subproblems, where each client $i$ solves its own subproblem with the objective of maximizing its reward as follows:
\begin{align}
&\max_{\theta_i, \varepsilon_i, \mathcal{C}_i} \: \mathcal{R}_i(h, \varepsilon, \mathcal{C}, \theta_g) - \mathrm{Cost}_i(\mathcal{C}_i) \\
&\quad\: \text{s.t.} \:\:\: h_i = \mathcal{M}_i(\theta_i, \varepsilon_i, \mathcal{C}_i), \nonumber \\
&\qquad\quad \varepsilon_{\text{L}} \leq \varepsilon_i \leq \varepsilon_{\text{U}}, \nonumber 
\end{align}
where $\theta_i$ corresponds to the parameters of client $i$, the activations $h_j$, $j \neq i$, the privacy budgets $\varepsilon_j$, $j \neq i$, the resource budgets $\mathcal{C}_j$, $j \neq i$, and the global model parameters $\theta_g$ are fixed to their values from the previous iteration, $\mathrm{Cost}_i$ corresponds to the cost incurred by client $i$, and $\mathcal{M}_i$ corresponds to the mapping between the parameters of client $i$ and its activation.  Client $i$ remains in the federation as long as its optimal objective value is greater than or equal to zero, ensuring that participation is individually rational and thus profitable.

\textbf{Problem Challenges}
Achieving the above objective involves addressing two key challenges:

\begin{itemize}
    \item \textit{Privacy-Utility Trade-off:} While differential privacy (DP) effectively protects individual data points from inference attacks, the injected noise can obscure important data patterns. This may lead to degraded global model accuracy. Achieving a balance between privacy and utility requires careful tuning of the noise to ensure sufficient protection without sacrificing predictive performance.
    
    \item \textit{Incentive-Aware Participation:} Clients, viewed as rational agents, assess their participation based on the trade-off between incurred costs (e.g., data transmission and computation) and expected rewards. If the perceived cost outweighs the benefit, clients may drop out, reducing both the quantity and quality of data contributions, which negatively impacts global model performance.
\end{itemize}
\texttt{OPUS-VFL} addresses these challenges as follows:
\begin{itemize}
    \item It uses a noise-aware reward mechanism that dynamically adjusts incentives based on clients' perturbed contributions, thus preserving privacy while promoting meaningful participation.
    
    \item It employs a leave-one-out (LOO) strategy to fairly assess client contributions under noisy conditions, ensuring reward allocation reflects true utility and encouraging continued participation.
\end{itemize}

\section{System Model} \label{sec:system-model}
In this section, we describe the structure and interactions of the VFL system in our \textit{network model}, highlighting collaborative training without raw data sharing. We then present the \textit{threat model}, which outlines privacy risks from \textit{honest-but-curious} (HBC) participants. Together, these models motivate the design and security considerations of the proposed \texttt{OPUS-VFL} framework.

\begin{SCfigure*}
    \centering
    \includegraphics[width=0.6\textwidth]{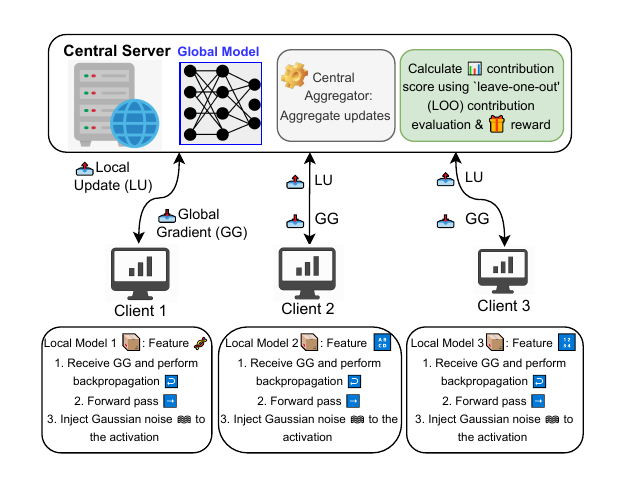}
    \caption{Overview of the OPUS-VFL framework: In this vertical federated learning (VFL) system, each client (e.g., Client~1, Client~2, and Client~3) holds a disjoint subset of the feature space and maintains a local model. During each training round, clients receive a global gradient (GG) from the central server and perform backpropagation locally. They then execute a forward pass and inject Gaussian noise into the activation to preserve privacy before sending the resulting local update (LU) to the central server. The central aggregator concatenates the received activations, computes the global loss and gradient, and updates the global model. It further evaluates each client’s contribution using a leave-one-out (LOO) strategy and distributes rewards accordingly. This framework enables privacy-preserving and incentive-aware collaborative learning without exposing raw feature data.}
    \label{fig:architecture}
\end{SCfigure*}

\subsection{Network Model}

We consider a vertical federated learning (VFL) environment consisting of a \textit{central server} and multiple \textit{clients} (e.g., Client~1, Client~2, and Client~3), each holding a disjoint subset of feature spaces corresponding to a shared sample ID space. The central server orchestrates the collaborative training of a \textit{global model}, while the clients maintain and update their respective \textit{local models} using private, non-overlapping features.

This setup adopts a \textit{SplitNN} architecture~\cite{liu2024vertical}, in which the server possesses labels, while each client holds only a subset of features. The system supports privacy-preserving, collaborative training without requiring clients to share raw data.

\subsubsection{\bf Client-Side Operations}
Each client is responsible for updating its local model using its private data. In each communication round, the following operations are performed:
\begin{enumerate}
    \item The client receives the \textit{Global Gradient (GG)} from the central server.
    \item It performs \textit{backpropagation} using the received gradient.
    \item The client executes a \textit{forward pass} on its local model.
    \item To enhance privacy using DP, \textit{Gaussian noise} is injected into the local activation before it is sent to the server.
\end{enumerate}

\subsubsection{\bf Server-Side Operations}
The central server coordinates the global learning process with the following steps:
\begin{enumerate}
    \item \textit{Aggregates local model updates} received from participating clients.
    \item Updates the \textit{global model} based on the aggregated information.
    \item \textit{Calculates contribution scores} for each client based on their model updates using LOO evaluation (see Section~\ref{subsec:client-opus-vfl})
    \item \textit{Distributes rewards} to the clients in proportion to their estimated contributions.
\end{enumerate}

The server initiates the training process and computes gradients based on the global loss, which are sent back to the clients for local model updates. To ensure efficient learning, the server aggregates client-side intermediate outputs via simple concatenation before computing the loss. Clients contribute to the training collaboratively by performing local forward passes and privately transmitting intermediate results without sharing raw features.

In \texttt{OPUS-VFL}, we consider one central server and $N \geq 2$ clients. This architecture enables \textit{privacy-preserving feature learning} through local noise injection and supports \textit{incentive-aware collaboration} via server-side reward allocation. The training protocol is executed iteratively, with gradient information and model updates exchanged between the server and clients to facilitate federated optimization.  \textbf{Figure~\ref{fig:architecture}} illustrates an overview of the \texttt{OPUS-VFL} framework.

\subsection{Threat Model} \label{subsec:threat-model}

We assume an \textit{honest-but-curious} adversarial model, in which both the server and clients adhere to the VFL protocol but may attempt to infer sensitive information from the shared intermediate representations. The threat surface in VFL includes several privacy risks arising from both the client and server sides.

Specifically, we consider the following attacks:
\begin{itemize}
\item \textit{Label inference attacks}: Malicious clients can craft adversarial updates or analyze gradient responses to infer label information from the server~\cite{fu2022label}. \texttt{OPUS-VFL} mitigates this risk by applying DP to the intermediate activations shared with the server, thereby obfuscating label-specific patterns in the gradients.

\item \textit{Feature inference attacks}: Malicious clients may attempt to reconstruct private features of other benign clients by analyzing the received intermediate representations~\cite{yang2023practical}. To defend against such leakage, \texttt{OPUS-VFL} adds Gaussian noise to the client-side activations before transmission. Importantly, the leave-one-out (LOO) contribution evaluation is performed directly on the perturbed activations, ensuring that raw feature embeddings are never revealed.

\item \textit{Backdoor attacks}: Malicious clients may inject poisoned samples or triggers into their local data to influence the behavior of the global model~\cite{naseri2024badvfl}. In \texttt{OPUS-VFL}, the addition of DP-based Gaussian noise in each training round helps suppress the influence of such backdoor triggers by reducing their detectability in the shared representations.
\end{itemize}
The \texttt{OPUS-VFL} framework is thus designed with built-in defense mechanisms, primarily through differentially private noise injection, that target and neutralize these key privacy threats while preserving the collaborative benefits of vertical federated learning.

\section{Proposed Approach: {\tt OPUS-VFL}}
\label{sec:opus-vfl-description}

This section outlines the core ideas of \texttt{OPUS-VFL}, our proposed approach to incentive-compatible and privacy-preserving vertical federated learning (VFL). We describe the overall training process, including how rewards are computed by the server, how clients adapt their behavior based on feedback, and how tokenized rewards are distributed to foster continued participation.

In \texttt{OPUS-VFL}, each client locally perturbs its model updates before transmitting them to the server, with the noise level calibrated to its chosen privacy budget. The server, which possesses critical features unavailable to clients, initiates the federated learning process but does not apply noise to its own updates. 

To optimize privacy settings dynamically, clients rely on feedback derived from rewards assigned at each training epoch. These rewards reflect the client’s contribution to the global model. Using backpropagation, we compute the gradient of each client's reward with respect to its privacy parameter, $\varepsilon$, enabling gradient-based tuning of noise levels. This optimization process allows clients to balance privacy preservation with their impact on the global model. Details on how the server computes feature importance to support this mechanism are provided in Section~\ref{subsubsec:server-feature-importance-derivation}.

\texttt{OPUS-VFL} integrates a reward-based incentive mechanism to encourage high-quality, privacy-aware contributions. Each client is modeled as a rational agent aiming to maximize its utility, quantified in reward tokens. These tokens may be exchanged for monetary incentives or computational privileges (e.g., bandwidth). The server provides per-round feedback on each client’s impact, enabling adaptive adjustment of privacy budgets and training strategies. Based on estimated utility, formalized in Eq.~\eqref{eq:legitimate-utility}, clients decide whether to continue participating in the VFL process.

The detailed procedures for both client- and server-side operations are presented in Sections~\ref{subsec:client-opus-vfl} and~\ref{subsec:server-opus-vfl}, respectively.

\subsection{Client-Side Incentive and Optimization Mechanisms}\label{subsec:client-opus-vfl}

\subsubsection{\bf Client's Feature Importance based on a Leave-One-Out (LOO) Strategy} 
\label{subsec:LOO}
The server evaluates the feature importance $\mathcal{I}_i$ of each client $i$ at the end of every training round by assessing the accuracy improvement attributed to its features. 
Specifically, the server compares the global model loss with and without client $i$'s features. 
A significant increase in loss upon removal indicates a higher feature contribution.

Let $h_i$ represent the local updates (activations) sent by client $i$ to the central server, and let the total number of clients be $N$. 
The server receives the set of updates (activations) $H = \{h_{1}, h_{2}, \ldots, h_{N}\}$ at the end of each training round and has access to the ground truth labels $y$.  To quantify the performance contribution of client $i \in \{1,\dots, N\}$, we use a leave-one-out (LOO) strategy to assess the importance of the feature vector it provides.

Since maximizing the global model's prediction accuracy is equivalent to minimizing its loss, we define the cross-entropy loss as:
\begin{equation}
\label{eq:cross-entr-loss}
\mathcal{L} = -\sum_{j=1}^{m} \left( y_j \log(p_j) + (1 - y_j) \log(1 - p_j) \right),
\end{equation}
where $m$ is the number of data samples, $y_j$ is the true label of sample $j$, and $p_j$ is the predicted probability of $j$ belonging to the positive class. 
The prediction $p_j$ is computed using all updates (activations) contributed by participating clients.

To measure client $i$'s feature importance $\mathcal{I}_i$, we define:
\begin{equation}
\label{eq:performanceContribution}
\mathcal{I}_i = \frac{(\mathrm{loss}_{N-i} - \mathrm{loss}_{N}) \times 100}{\mathrm{loss}_{N}},
\end{equation}
where the loss values are:
\begin{eqnarray}
\label{eq:loss-N-i}
\text{loss}_{N-i} &=& \mathcal{L}(\mathcal{M}(\mathcal{H} - \{h_{i}\}), y),  \\    
\text{loss}_{N} &=& \mathcal{L}(\mathcal{M}(\mathcal{H}), y). \nonumber
\end{eqnarray}
Here, $\text{loss}_{N-i}$ represents the model's loss when client $i$'s updates (activations) are excluded, while $\text{loss}_{N}$ corresponds to the loss when all clients' updates are included. 
A significant increase in loss upon removing client $i$'s updates indicates its contributions are critical for minimizing the global model loss.
To estimate $\text{loss}_{N-i}$, we train a separate model that excludes client $i$'s updates (activations) during each training round. 
A higher value of $\mathcal{I}_i$ signifies greater importance of client~$i$'s features in reducing the global model loss.

Since contribution assessment requires an optimized model, each client's local model must be trained under its selected privacy budget before its importance can be computed. To reduce the computational overhead of collaboratively retraining the full VFL model, client contributions are evaluated only after a fixed number of initial epochs.

\subsubsection{\bf Client Objective}
Each client's primary objective is to maximize its utility while balancing contribution quality and privacy budget. 
Our system rewards clients who strike this balance by dynamically adjusting their privacy budgets in each training round. 
Clients have the flexibility to choose their privacy levels while ensuring their utility remains high. 

The privacy term is designed as a function of both $\Delta f_i$ (sensitivity) and $\varepsilon_{it}$ (privacy budget), where higher privacy corresponds to a larger privacy budget. 
Thus, each client optimizes two competing objectives. 
At each training round $t$, the objective function for client $i$ is given by:
\begin{align}
\label{eq:reward}
\max_{\varepsilon_{it} \in [\varepsilon_\text{L}, \varepsilon_{\text{U}}]} \mathcal{R}_{it} = \alpha \cdot \mathcal{S}_{it} + \beta \cdot \mathcal{P}_{it},
\end{align}
where $\varepsilon_{\text{L}}$ and $\varepsilon_{\text{U}}$ denote the lower and upper bounds of the privacy parameter $\varepsilon_{it}$, and $\alpha$ and $\beta$ are tuning parameters that balance contribution and privacy (see Section~\ref{subsec:design-parameter}).  The individual components of the objective function are:
\begin{equation}
\mathcal{S}_{it} = \mathcal{I}_{it}\mathcal{C}_{it}^{1/a}, \; \; 
\mathcal{P}_{it} = \frac{\Delta f_i}{\varepsilon_{it}},
\end{equation}
where $\mathcal{S}_{it}$ represents the contribution-related term, $\mathcal{P}_{it}$ represents the privacy-related term, $\mathcal{I}_{it}$ denotes the client's feature importance (determined by the server), and $\mathcal{C}_{it}$ is the fraction of the cost incurred by client $i$ relative to its total available training budget in round $t$. 
The parameter $a$ (chosen by the server) determines the emphasis on equity in clients' resource availability.

Clients optimize $\varepsilon_{it}$ in each training round using feedback from the gradient of the objective function (Eq.~\eqref{eq:reward}) with respect to $\varepsilon_{it}$ (Eq.~\eqref{eq:partial-derivatives-chain-rule}). Training costs incurred by clients are also incorporated into the reward, ensuring fair compensation for those contributing maximally within their available resources.

Our approach incentivizes clients to maximize their contribution within their computational limits, preserving both privacy and global model accuracy.  The parameter $a$ allows the server to adjust the importance of opportunity fairness, controlling how the clients' utilized resources impact the reward calculation. 
We provide further details on this parameter in Section~\ref{sec:exp-setup}.

\subsubsection{\bf Utility Function of a Legitimate Client}
A legitimate client, one that does not engage in any form of attack, participates in VFL as a rational entity aiming to maximize its utility. This utility is defined by the trade-off between participation costs (e.g., energy or computational resources) and the expected rewards issued by the server.  The utility of a legitimate client in training round~$t$, denoted as $\mathcal{U}_{it}$, is given by:
\begin{eqnarray}
\label{eq:legitimate-utility}
\mathcal{U}_{it} = \tau_{it} - \mathcal{B}_{it},
\end{eqnarray}
where $\tau_{it}$ denotes the number of reward tokens received by the client, computed according to the procedure in \textbf{Algorithm~\ref{alg:rewards_distribution}}, and $\mathcal{B}_{it}$ represents the actual cost incurred for training the local model in round~$t$.  This utility serves as the basis for a client to decide whether to participate in the FL process.  In this work, a client withdraws from the FL process when its utility falls below zero.

\subsubsection{\bf Client's Gaussian Noise Computation to Enhance Privacy Using DP}  To ensure differential privacy (DP), each client $i$ adds Gaussian noise to its local model updates.  The noise is generated by drawing a random number for each element of a data point vector $j$ and scaling it based on the variance of the Gaussian distribution.   In each training round, client $i$ applies the Gaussian mechanism by:
\begin{gather}
\mathcal{N}(h_{it}, f, \varepsilon_{it}, \delta_i) = h_{it} + \mathcal{N}_N \left(\mu=0, \sigma^2=\frac{2\ln(1.25/\delta_i)\times \Delta f^2}{\varepsilon_{it}^2} \right).
\end{gather}
where $h_{it}$ is the original value to be added noise to, $f$ is the function with sensitivity $\Delta f$, and $\varepsilon_{it}$ and $\delta_i$ are the privacy parameters governing the noise magnitude. The added noise is drawn from a Gaussian distribution $\mathcal{N}_N(0, \sigma^2)$ with variance calibrated to ensure ($\varepsilon_{it}$, $\delta_i$)-differential privacy.
The perturbed update vector $h_{it}$ is computed as follows. Let $r_{it}$ denote a random vector sampled by client~$i$, where each element corresponds to an independently drawn random value. Each element of the local model update vector is then perturbed using the corresponding element in $r_{it}$.  The gradient of $h_{it}$ with respect to $\varepsilon_{it}$ is computed as:
\begin{eqnarray}
\label{eq:gaussian-noise}
\frac{\partial h_{it}}{\partial \varepsilon_i} &=& r_{it} \times \frac{\partial}{\partial \varepsilon_i} \sqrt{\frac{2\ln(1.25/\delta_i)\times \Delta f^2}{\varepsilon_{it}^2}}  \\    
&=& -r_{it} \times \frac{\sqrt{2\ln(1.25/\delta_i) \times \Delta f^2}}{\varepsilon_{it}^2}. \nonumber
\end{eqnarray}

\subsection{Server-Side Feedback Computation and Client Evaluation} \label{subsec:server-opus-vfl}
\subsubsection{\bf The Server's Feature Importance Derivation} \label{subsubsec:server-feature-importance-derivation}
The reward assigned to client $i$ depends on its update $h_i$, which is influenced by the privacy parameters $(\varepsilon_{it}, \delta_i)$. To determine the optimal $\varepsilon_{it}$ for the next training round, we apply the chain rule to compute the gradient of $\mathcal{I}_{it}$ with respect to $\varepsilon_{it}$ as follows:
\begin{equation}
\label{eq:partial-derivatives-chain-rule}
G_{\mathrm{contribution}} = \alpha \left(\frac{\partial \mathcal{S}_{it}}{\partial h_{it}} \cdot \frac{\partial h_{it}}{\partial \varepsilon_{it}}\right) + \beta \frac{\partial \mathcal{P}_{it}}{\partial \varepsilon_{it}},
\end{equation}
where $\alpha$ and $\beta$ are tuning parameters that balance contribution and privacy (see Section~\ref{subsec:design-parameter}). Using this gradient, the privacy parameter $\varepsilon_i$ for client $i$ in the next training round is updated as:
\begin{equation}
\label{eq:epsilon-maximization}
\varepsilon^{r+1}_{it} = \min\Big\{ \varepsilon_{\text{U}}, \max \Big\{ \varepsilon_{\text{L}}, \varepsilon^{r}_{it}+G_{\mathrm{contribution}}\Big |_{\varepsilon^{it} = \varepsilon^r_{it}} \Big\} \Big\}.
\end{equation}

\subsubsection{\bf Rewards Distribution by the Server}
After computing rewards using Eq.~\eqref{eq:reward}, the server distributes them to maintain budget balance while enabling seamless exchange for monetary benefits or computational resources. 
These reward tokens can be traded, redeemed, or allocated based on clients' contributions and needs. 
The distribution process is detailed in \textbf{Algorithm~\ref{alg:rewards_distribution}}.

The rewards distribution algorithm allocates rewards to clients based on their contributions and privacy preferences during training. 
It takes as input the computed rewards $\mathcal{R}$ from Eq.\ \eqref{eq:reward} and the total number of reward tokens per round ($\tau_{ar}$), a fixed budget assigned by the server as the initiator of the VFL process.
The reward allocation $\tau_i$ for each client $i$ in the current round is determined based on \textbf{Algorithm~\ref{alg:rewards_distribution}}. 
The assigned tokens $\tau_i$ are then deducted from the remaining budget~$\tau_{ar}$.  If tokens remain ($\tau_{ar} > 0$), they are distributed to clients using a round-robin allocation strategy. However, if no tokens are left, clients that did not receive any allocation are permanently dropped from the federation.  This process is repeated for all clients throughout training rounds until model training is complete.

\begin{algorithm}[tb]
\caption{Rewards Distribution}
\label{alg:rewards_distribution}
\begin{algorithmic}[1]
  \State \textbf{Input:} Rewards $\mathcal{R}$, number of reward tokens $\tau_{ar}$ per round, number of clients $N$, number of training rounds $T$
  
  \For{$t = 1$ to $T$}
    \For{$i = 1$ to $N$}
      \State \textbf{Compute the reward token distribution} $\tau_{it}$:
      \State
      \[
        \tau_{it} = 
          \left(\frac{\mathcal{R}_{it}}{\sum_{j=1}^{N} \mathcal{R}_{jt}}\right)
            \times 
          \left(\frac{N \times (N+1)}{2}\right)
      \]

      \State \textbf{Update the remaining tokens}:
      \State
      \[
        \tau_{ar} \gets \tau_{ar} - \tau_{it}
      \]

      \If{$\tau_{ar} > 0$}
        \State Assign remaining tokens to all clients in a round-robin fashion.
      \EndIf
    \EndFor
  \EndFor
  \State \textbf{return} $\tau$
\end{algorithmic}
\end{algorithm}

\subsubsection{\bf Handling Drop-Out Scenarios}
\texttt{OPUS-VFL} undergoes a warm-up phase for a predefined number of epochs before distributing incentives.  During this phase, we assume that all clients remain in the federation.  At the end of the warm-up period, clients are given the option to drop out if their utility falls below their expectations.  Similarly, the server reserves the right to remove clients whose contributions are insufficient.  Once the warm-up phase concludes, all remaining clients continue participating in the federation, as their utility is ensured to be above a predefined threshold (e.g., 0).

\section{Experiment Setup} \label{sec:exp-setup}
\subsection{Design Parameters} \label{subsec:design-parameter}

\begin{table}[t]
\centering
\caption{\centering Design Parameters and Their Lower/Upper Bounds Used in \texttt{OPUS-VFL}} \label{tab:design-choices}
\begin{tabular}{ccc}
\toprule
Design parameter & Lower bound & Upper bound \\ \midrule
$\varepsilon$    & 0.5         & 5           \\
$\Delta f$        & 0.01        & 1           \\
$a$              & 2           & 5           \\
$\tau_{ar}$      & 10         &N/A             \\
$N$   & 2            & 25           \\
$\alpha$, $\beta$   & 0.1            & 1           \\
$\mathcal{C}$   & $0.01$            & 1         \\
$\mathcal{B}$   & 50 & N/A\\
\bottomrule
\end{tabular}

(Note: The privacy parameter $\varepsilon$ can be tuned in every round, while the other parameters are fixed at the beginning of training.)
\end{table}

\textbf{Table~\ref{tab:design-choices}} presents the upper and lower bounds for the tunable design parameters used in \texttt{OPUS-VFL}. For non-tunable parameters such as $\mathcal{L}$, $\mathcal{U}$, and $\mathcal{R}$, their bounds depend on the range of $\mathcal{I}$. The upper bound on the loss $\mathcal{L}$ can be estimated by computing the loss from the initial model before training. Although the theoretical lower bound of loss is zero, in practice, each model may have a different lower bound depending on the dataset used.

To ensure all parameters lie within a comparable numerical range, we set the scaling parameters $\alpha$ and $\beta$ so that $\mathcal{S}$ and $\mathcal{P}$ (defined in Eq.~\eqref{eq:reward}) are appropriately normalized. In our MNIST experiments, we found $\alpha = 1$ and $\beta = 10$ to be suitable.  We chose bounds for $\varepsilon$ and $\Delta f$ such that changes in these values do not significantly degrade prediction accuracy. For example, if a client selects a very small privacy budget (e.g., $\varepsilon = 0.01$) to preserve privacy, the resulting noise in its model updates may adversely affect the training of other clients in the federation. We set the bounds to avoid such situations. To ensure sufficient noise while preserving utility, we evaluated the signal-to-noise ratio (SNR) for various combinations of $\varepsilon$ and $\Delta f$.

Each client can configure the parameter $a$ based on the level of opportunity fairness they wish to consider—specifically, the weight assigned to the fraction of computational resources used in the reward computation. By setting the lower bound of $\mathcal{C}$ to 0.01, we require clients to contribute at least 1\% of their available resources for training. This parameter can also be adapted based on the types of clients the VFL system aims to accommodate. For example, one client may have only a personal computer but hold valuable data, while another might be a large tech company with abundant resources. If both types participate in \texttt{OPUS-VFL}, the range of $\mathcal{C}$ must be broad enough to reflect disparities in available resources—for instance, from $10^{-5}$ to $100$. Despite such variation, each client can still receive fair incentives by appropriately choosing the parameter $a$. The bounds listed in \textbf{Table~\ref{tab:design-choices}} assume clients are mid-sized organizations with meaningful data.

Finally, the parameter $\tau_{ar}$ is set to ensure no client drops out during training. While the system could potentially fail if $\mathcal{B} > \text{contributions}$, we reasonably assume that clients incurring higher costs are also making more significant contributions, i.e., $\mathcal{B} < \text{contributions}$ in most practical scenarios.

\subsection{Datasets}

We evaluate our approach using the following datasets:
\begin{itemize}
  \item \textbf{MNIST}: This dataset consists of 70,000 grayscale images of handwritten digits ($28 \times 28$ pixels) across 10 classes. We include MNIST due to its simple, tabular-like structure, which facilitates straightforward analysis of model behavior.

  \item \textbf{CIFAR-10}: This dataset includes 60,000 training and 10,000 test RGB images ($32 \times 32$ pixels) across 10 object categories. Widely used as a benchmark for image classification, it helps assess our method's effectiveness on a standard, moderately complex task.

  \item \textbf{CIFAR-100}: CIFAR-100 extends CIFAR-10 with images from 100 fine-grained object classes, presenting a more challenging classification task. We use it to evaluate the robustness and scalability of our approach in high-class-count scenarios.
\end{itemize}

\subsection{Model Architectures}
\textbf{Table~\ref{tab:model_architectures}} summarizes the model architectures used for each dataset in our experiments. To ensure fair comparisons, we applied these architectures consistently across all baseline methods when evaluating the resilience of \texttt{OPUS-VFL} against various attacks.

\begin{table}[t]
\centering
\caption{Model Architectures}
\label{tab:model_architectures}
\begin{tabular}{ccc}
\toprule
\textbf{Dataset} & \textbf{Clients} & \textbf{Server} \\
\midrule
CIFAR-10        & ResNet-18 & FCNN-3 \\
CIFAR-100       & ResNet-18 & FCNN-3 \\
MNIST        & FCNN-2 & FCNN-1 \\
\bottomrule
\end{tabular}

(Note: ``FCNN-3'' refers to the 3-layer \\ fully connected neural network.)
\end{table}

\subsection{Hyperparameters for VFL} We use a shared set of client and server learning rates for vanilla VFL and incentive-based baselines, and a slightly different set for privacy-focused ones to better align attack and model accuracy. \texttt{OPUS-VFL} is evaluated under multiple attack scenarios, with consistent hyperparameters across different attack strengths. For DP-based baselines, including \texttt{OPUS-VFL}, we slightly adjust learning rates to maintain comparable performance. This approach is also used when replicating baseline implementations from~\cite{naseri2024badvfl, fu2022label} for backdoor and label inference evaluations.

The definition of attack strength varies by type and is detailed in Section~\ref{sec:p2vfl-against-attacks}. All \texttt{OPUS-VFL} experiments were run for different numbers of training epochs depending on the dataset. To ensure consistency and stability, adversarial robustness and incentive mechanism evaluations (i.e., contribution, reward, and equity) were conducted over 10 simulation iterations each.

\subsection{Comparison Schemes}

The following schemes are considered for performance comparison:
\begin{itemize}
\item \textbf{Vanilla VFL Model}~\cite{vepakomma2018split}: A basic VFL setup with a central trainable server and $N \geq 2$ clients. This model does not incorporate any privacy guarantees or contribution evaluation mechanisms.

\item \textbf{Theoretically Optimal Organization Selection Mechanism (O2S)}~\cite{tan2023fraim}: O2S computes an organization’s importance by summing its feature importance values, and a coordinator performs a first-depth search to identify the optimal organization combination. However, O2S does not guarantee truthful reporting from participants.

\item \textbf{Bid Price First (BPF) Mechanism}~\cite{tan2023fraim}: In this scheme, the server selects clients in ascending order of reported costs. Like O2S, BPF does not guarantee truthful reporting.

\item \textbf{Privacy-Preserving Deep Learning (PPDL)}~\cite{shokri2015privacy}: PPDL is a widely used gradient-level defense mechanism in deep learning. It integrates differential privacy, gradient sparsification, and randomized selection. Specifically, in each training iteration, gradient components are randomly selected and perturbed with Gaussian noise. Only those exceeding a threshold $\tau$ in magnitude are retained, and this process continues until a target fraction $\theta_u$ of gradients is preserved. This approach balances privacy and model performance and is integrated into our vertical federated learning pipeline. PPDL is noted as a defense against label inference attacks in~\cite{fu2022label}. We include PPDL in our comparison as its threat model aligns closely with that addressed by \texttt{OPUS-VFL}.

\item \textbf{VFL-CZOFO}~\cite{wang2024unified}: VFL-CZOFO is a unified optimization framework that enhances privacy and communication efficiency in VFL. It employs Zero-Order Optimization (ZOO) on the client's output layer to ensure privacy, and First-Order Optimization (FO) for other layers to improve training performance. ZOO gradients are estimated via random perturbations and offer implicit $(\varepsilon, \delta)$-differential privacy guarantees.
\end{itemize}

\subsection{Metrics} We evaluate all schemes using the following metrics:
\begin{itemize}
    \item \textbf{Prediction Accuracy} ($\mathcal{A}$): Measures the global model’s accuracy as the proportion of correct predictions among all predictions, defined as:
    \begin{equation}
        \mathcal{A} = \frac{\sum_{i=1}^m \mathbb{1}(\hat{y}_i = y_i)}{m}
    \end{equation}
    where $m$ is the total number of samples, $\hat{y}_i$ is the predicted label for the $i$-th sample, $y_i$ is the true label, and $\mathbb{1}(\cdot)$ is the indicator function that returns 1 if the condition is true and 0 otherwise.

    \item \textbf{Mean Training Time to Convergence} ($\mathcal{T}_{\text{conv}}$): Measures the average time required to complete a single training round on one batch of data, used to evaluate training efficiency.

    \item \textbf{Attack Success Rate (ASR)}: Measures the effectiveness of a backdoor attack in compromising the VFL model. It is defined as the percentage of successful attack attempts:
    \begin{equation}
        \text{ASR} = \frac{x_s}{x_t} \times 100\%,
    \end{equation}
    where $ x_s $ is the number of successful attacks, and $ x_t $ is the total number of attack attempts. For example, if 30 out of 100 adversarial inputs succeed, the ASR is 30\%. A higher ASR indicates a more effective attack.

    \item \textbf{Reconstruction Accuracy (RA)}: Assesses the effectiveness of inference attacks to reconstruct sensitive labels or features. It is measured using the Mean Squared Error (MSE) between true and reconstructed values.

    For labels, we measure $\text{ACC}_{\text{label}}$ by:
    \begin{equation}
        \text{ACC}_{\text{label}} = \frac{1}{D} \sum_{i=1}^{D} \left( y_i - \hat{y}_i \right).
    \end{equation}

    For features, we measure $\text{MSE}_{\text{feature}}$ by:
    \begin{equation}
        \text{MSE}_{\text{feature}} = \frac{1}{D} \sum_{i=1}^{D} \left( x_i - \hat{x}_i \right)^2.
    \end{equation}
    where $ D $ is the number of samples; $ y_i $ (or $ x_i $) are the ground-truth labels (or features); and $ \hat{y}_i $ (or $ \hat{x}_i $) are the reconstructed counterparts. A lower MSE implies better reconstruction accuracy, indicating more successful inference by the adversary.
\end{itemize}

\begin{table}[t]
\centering
\caption{\centering Performance Comparison Across VFL Baselines on CIFAR-10 (5 Clients)}
\label{tab:baseline-comparison-cifar10}
\begin{tabular}{@{}lcc@{}}
\toprule
\textbf{Method} & \textbf{Accuracy} & \textbf{Avg. Time (s)} \\
\midrule
Vanilla VFL     & 62.9\%  & 5.2  \\
O2S             & 63.2\%  & 6.0  \\
BPF             & 61.3\%  & 4.8  \\
VFL-CZOFO       & 53.8\%  & 32.2 \\
VFL-SGD         & 57.0\%  & 9.0  \\
\texttt{OPUS-VFL} & 60.4\%  & 3.8  \\
\bottomrule
\end{tabular}
\end{table}

\section{Experimental Results \& Analyses} \label{sec:results-analysis}

This section presents empirical results demonstrating the efficiency, scalability, and robustness of \texttt{OPUS-VFL}. Evaluations on CIFAR-10, CIFAR-100, and MNIST show that \texttt{OPUS-VFL} delivers strong performance, efficient training, and enhanced privacy protection compared to state-of-the-art baselines.

\subsection{Training Efficiency and Scalability of \texttt{OPUS-VFL}} \label{subsec:scalability-analysis}

\textbf{Table~\ref{tab:baseline-comparison-cifar10}} provides a comprehensive comparison of the average per-round training time across various baselines on the CIFAR-10 dataset with 5 clients, along with a scalability analysis of \texttt{OPUS-VFL} on the MNIST dataset as the number of clients increases.  \texttt{OPUS-VFL} achieves the lowest average per-round training time of 3.8 seconds, outperforming all other methods. Compared to VFL-CZOFO, the slowest baseline at 32.2 seconds per round, \texttt{OPUS-VFL} is over 88\% more efficient. It also improves upon Vanilla VFL (5.2s) and VFL-SGD (9.0s) by approximately 26.9\% and 57.8\%, respectively. Even relative to BPF, the fastest among the other baselines (4.8s), \texttt{OPUS-VFL} achieves a 20.8\% reduction in computation time.

\begin{table}[t]
\centering
\caption{\centering Scalability of \texttt{OPUS-VFL} on MNIST With Varying Numbers of Clients}
\label{tab:scalability-opusvfl-mnist}
\begin{tabular}{@{}ccc@{}}
\toprule
\textbf{\# Clients} & \textbf{Accuracy} & \textbf{Avg. Time (s)} \\
\midrule
5   & 88.65\% & 3.82  \\
10  & 92.50\% & 6.20  \\
15  & 91.25\% & 8.51  \\
20  & 91.25\% & 12.35 \\
25  & 88.75\% & 16.23 \\
\bottomrule
\end{tabular}
\end{table}

\textbf{Table~\ref{tab:scalability-opusvfl-mnist}} demonstrates the scalability of \texttt{OPUS-VFL} on MNIST. As the number of clients increases from 5 to 25, the global model maintains competitive accuracy, peaking at 92.5\% with 10 clients, while the average training time per epoch naturally increases. These results highlight \texttt{OPUS-VFL}'s strong scalability, achieving efficient training without significant degradation in model performance, thereby reinforcing its practical utility for large-scale VFL systems.

In our scalability experiments, we assessed \texttt{OPUS-VFL} under increasing numbers of clients to evaluate both training efficiency and scalability. While several privacy-preserving VFL baselines exist, they were excluded from these experiments due to practical scalability limitations. For instance, FRAIM~\cite{tan2023fraim} involves Shapley value computation, which is computationally intensive, requiring over 1.5 hours even for just 7 clients on MNIST, making it impractical for larger-scale evaluation. Similarly, FedSDG-FS++~\cite{li2023fedsdg} relies on homomorphic encryption, significantly increasing runtime and limiting its suitability for adversarial robustness studies.

We also excluded TEA~\cite{lu2022truthful} due to the lack of an official implementation. Moreover, extending FRAIM to CIFAR-10 was infeasible due to the complexity of Hermite extrapolation in high-dimensional feature spaces. Although these methods offer strong theoretical guarantees, their high computational costs highlight limited practical scalability. In contrast, \texttt{OPUS-VFL} stands out as a lightweight and efficient alternative, well-suited for realistic and large-scale federated learning scenarios.

\subsection{Contribution, Reward, and Privacy Dynamics on Training Behavior in \texttt{OPUS-VFL}}

\begin{figure*}
\centering
\begin{subfigure}[b]{0.3\textwidth}
\centering
\includegraphics[width=\textwidth]{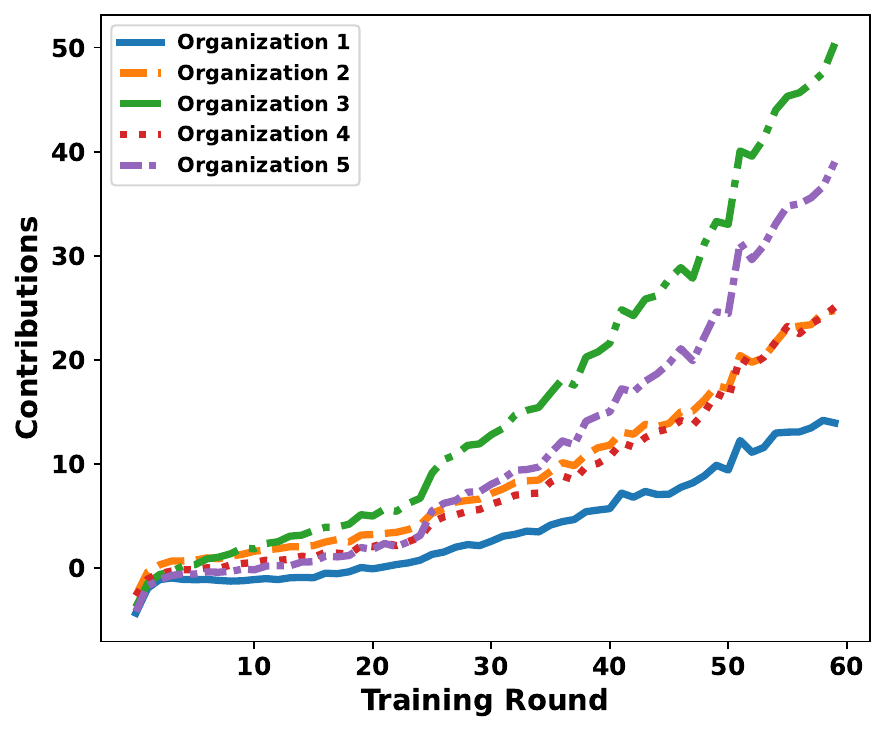}
\caption{Contributions}
\end{subfigure}
\begin{subfigure}[b]{0.3\textwidth}
\centering
\includegraphics[width=\textwidth]{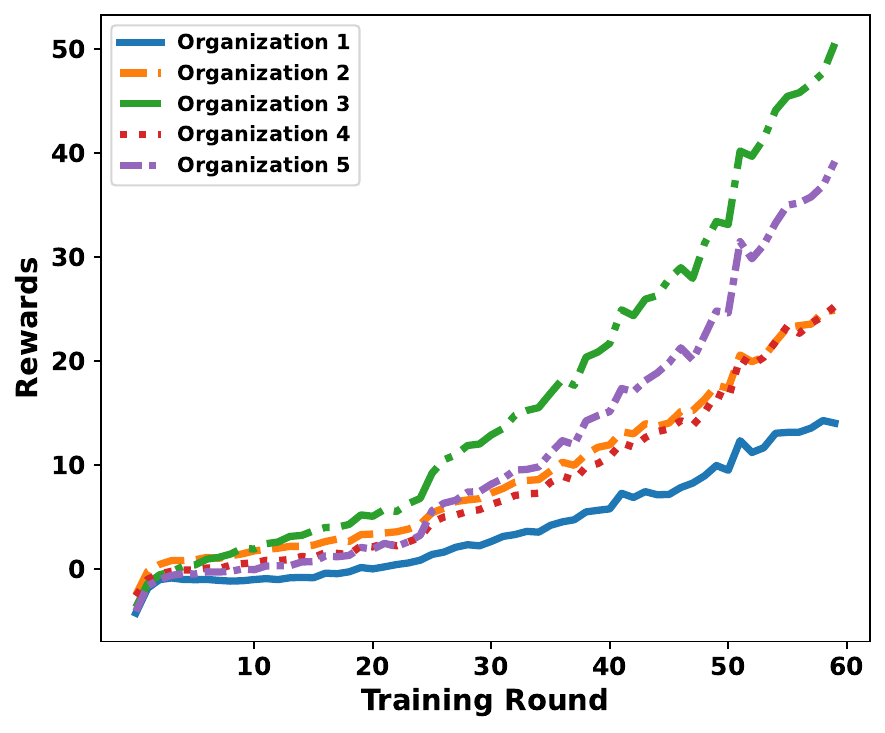}
\caption{Rewards}
\end{subfigure}
\begin{subfigure}[b]{0.3\textwidth}
\centering
\includegraphics[width=\textwidth]{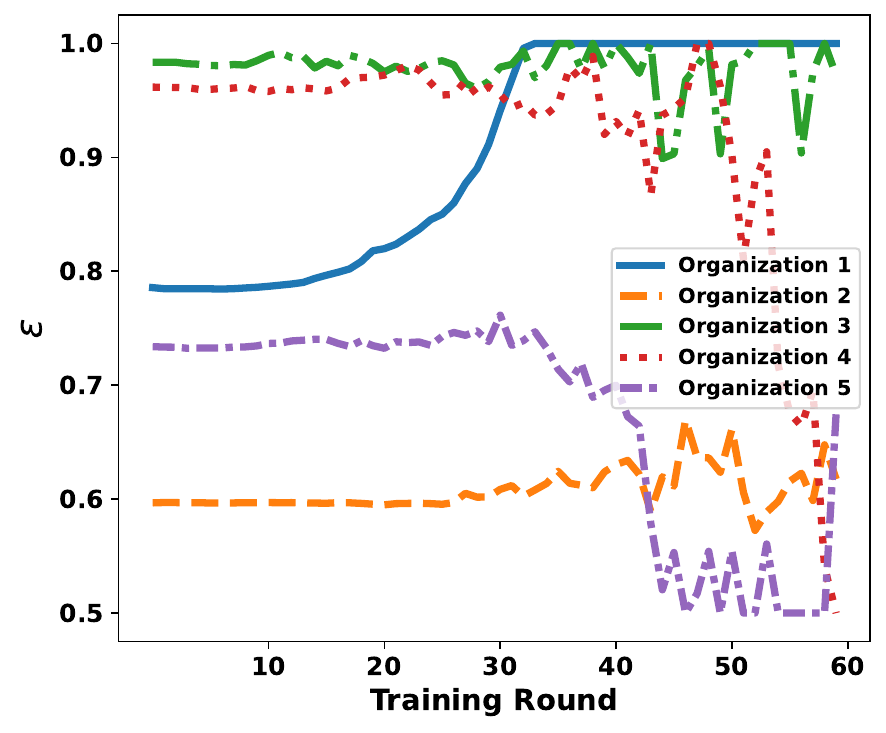}
\caption{Adopted $\varepsilon$ (DP parameter)}
\end{subfigure}
\caption{\textbf{Evolution of Contributions, Rewards, and Differential Privacy Parameter $\varepsilon$ Values:} Contributions, rewards, and differential privacy parameter $\varepsilon$ values generated by \texttt{OPUS-VFL} on the CIFAR-10 dataset over 60 training rounds. The $\varepsilon$ values remain relatively stable across epochs, resulting in similar trends for rewards and contributions. The magnitude of change in $\varepsilon$ can be controlled via a step size parameter.} 
\Description{OPUS-VFL training plots}
\label{fig:cifar10-plots}
\end{figure*}

\begin{figure*}
\centering
\begin{subfigure}[b]{0.3\textwidth}
\centering
\includegraphics[width=\textwidth]{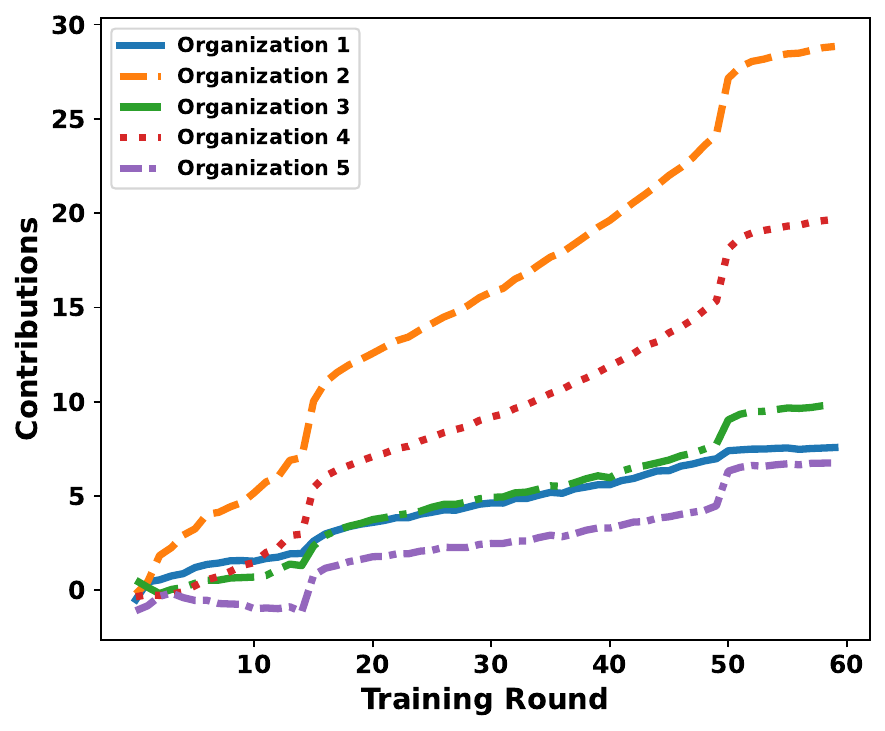}
\caption{Contributions}
\end{subfigure}
\begin{subfigure}[b]{0.3\textwidth}
\centering
\includegraphics[width=\textwidth]{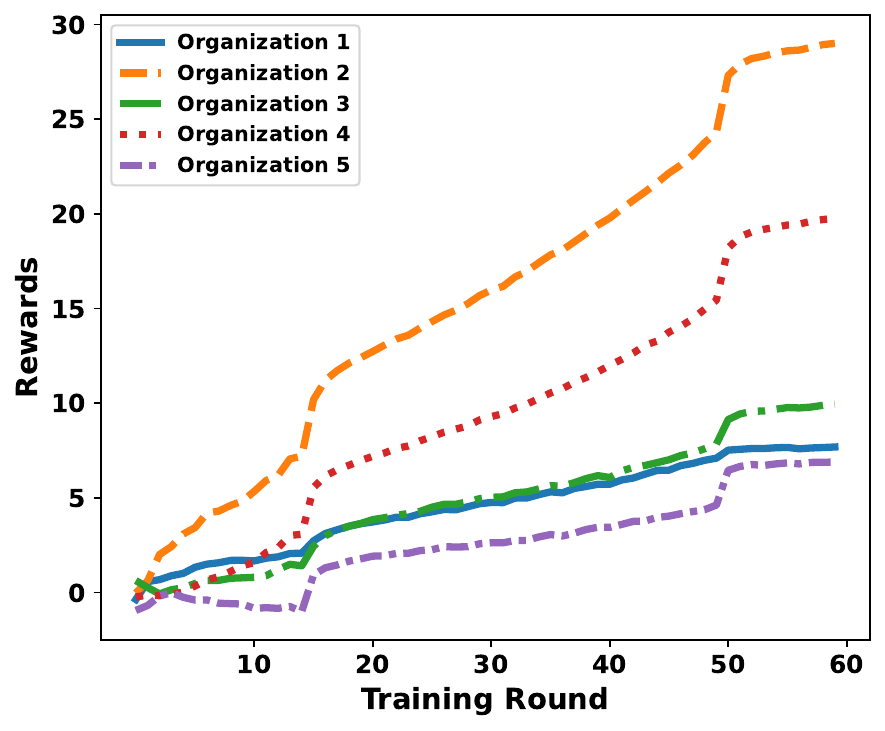}
\caption{Rewards}
\end{subfigure}
\begin{subfigure}[b]{0.3\textwidth}
\centering
\includegraphics[width=\textwidth]{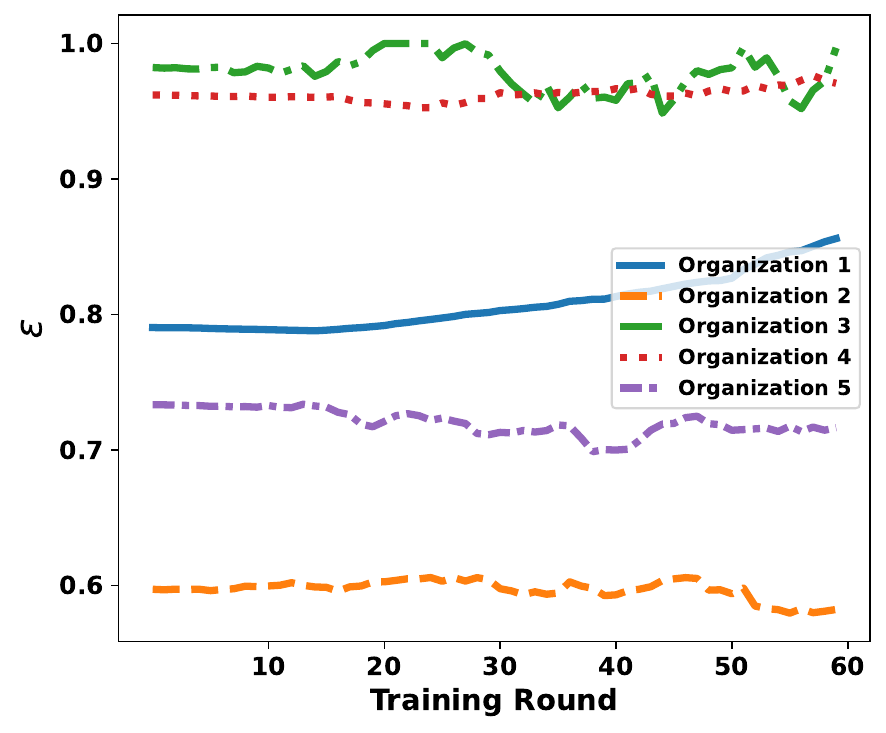}
\caption{Adopted $\varepsilon$ (DP parameter)}
\end{subfigure}

\caption{Evolution of Contributions, Rewards, and Privacy Parameter $\varepsilon$ for Each Client over 60 Epochs on CIFAR-100.}
\Description{OPUS-VFL contribution equity plots}
\label{fig:cifar100-plots}
\end{figure*}

In \textbf{Figures~\ref{fig:cifar10-plots} and ~\ref{fig:cifar100-plots}}, we illustrate how each participating organization's contribution, reward, and $\varepsilon$-values evolve during training on the CIFAR-10 dataset. The contribution curves (left panels) show how each party’s data and model updates enhance the global objective, with noticeable variation across organizations. Ensuring convergence of these updates is essential; otherwise, contribution measurements may be inaccurate. In practice, we observe that these curves stabilize in the later training rounds, aligning with the convergence of the global model.

The middle panels display the cumulative rewards accrued by each organization, representing the marginal utility attributed to their respective updates. While these reward trajectories often mirror the contribution curves, they may diverge when certain organizations contribute more informative feature subsets.

The $\varepsilon$-value plots (right panels) highlight the impact of privacy tuning on training dynamics. Overly aggressive tuning (e.g., setting $\varepsilon$ too high) can destabilize the training process and lead to divergence. In contrast, moderately sized $\varepsilon$-values provide a favorable trade-off between convergence speed and model stability. Importantly, when the dataset contains a sufficiently rich feature space, as is the case with CIFAR-10, \texttt{OPUS-VFL} demonstrates strong scalability across multiple organizations without sacrificing convergence or the accuracy of contribution estimation.

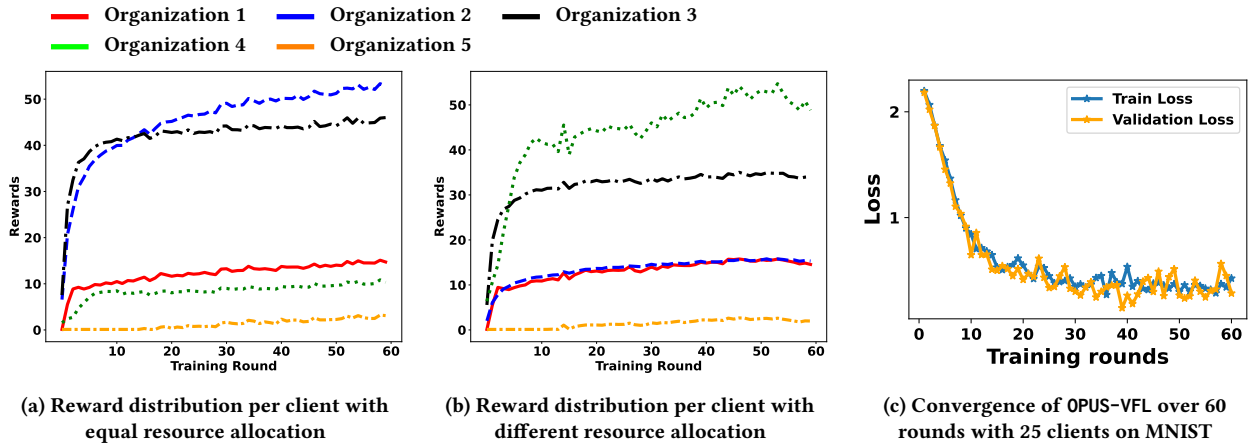
\begin{figure*}
\centering
\raggedright 
\hspace*{0.5cm}
  \begin{subfigure}[b]{0.6\textwidth}
    \begin{tikzpicture}[every node/.style={font=\small}]
      \foreach \x/\col/\name in {%
          0/red/Organization 1, 
          3/blue/Organization 2, 
          6/black/Organization 3%
      } {
          \draw[line width=2pt, \col] (\x,0) -- ++(0.5,0);
          \node[anchor=west] at ($(\x,0)+(0.6,0)$) {\textbf{\name}};
      }
    \end{tikzpicture}
  \end{subfigure}\\[-0.1cm]
\hspace*{0.5cm}
  \begin{subfigure}[b]{0.6\textwidth}
    \begin{tikzpicture}[every node/.style={font=\small}]
      \foreach \x/\col/\name in {%
          9/green/Organization 4, 
          12/orange/Organization 5%
      } {
          \draw[line width=2pt, \col] (\x,0) -- ++(0.5,0);
          \node[anchor=west] at ($(\x,0)+(0.6,0)$) {\textbf{\name}};
      }
    \end{tikzpicture}
  \end{subfigure}
\hfill\\
\begin{subfigure}[b]{0.30\textwidth}
\centering
\includegraphics[width=\textwidth]{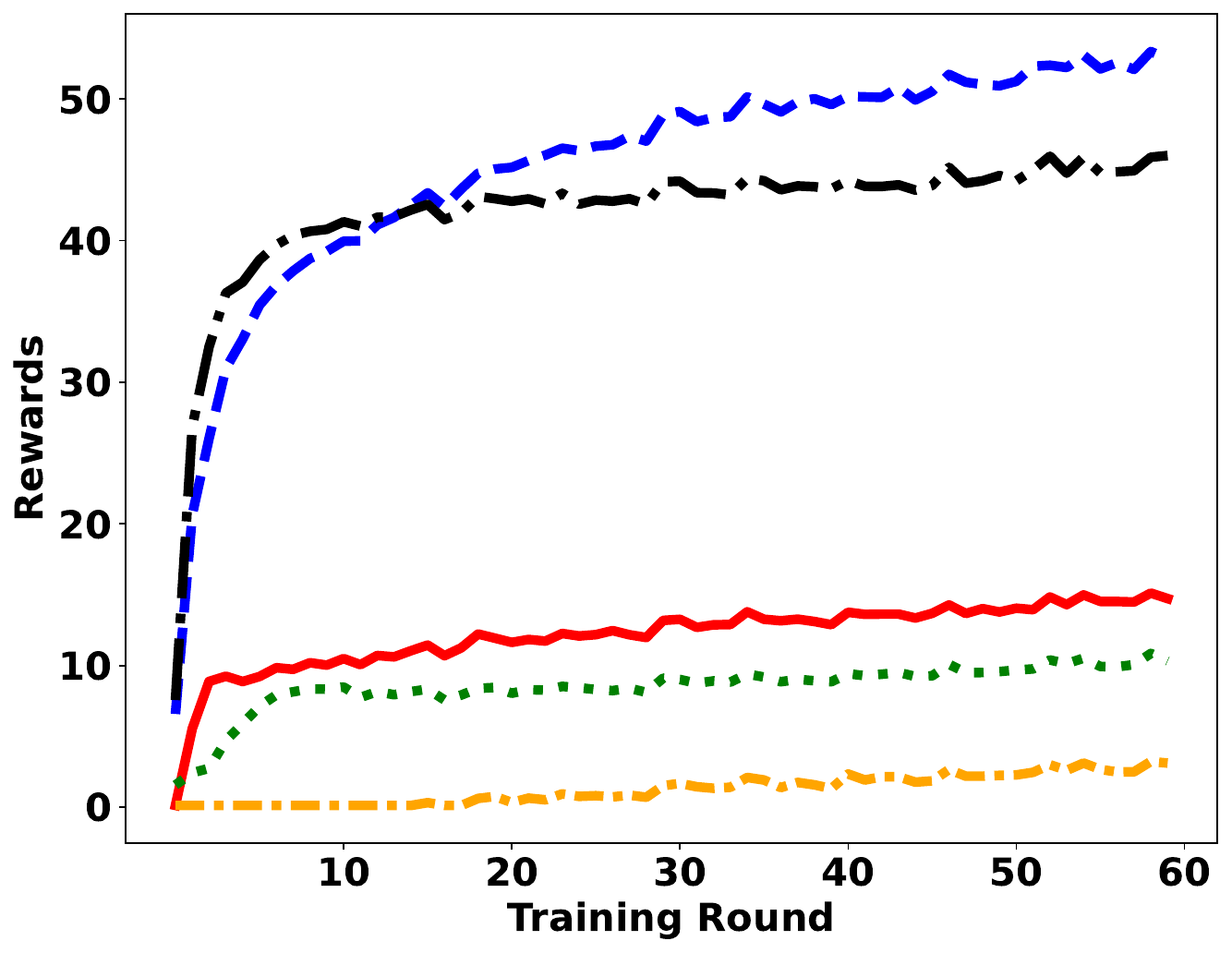}
\caption{\centering Reward distribution per client with equal resource allocation}
\end{subfigure}
\hspace{0.06in}
\begin{subfigure}[b]{0.30\textwidth}
\centering
\includegraphics[width=\textwidth]{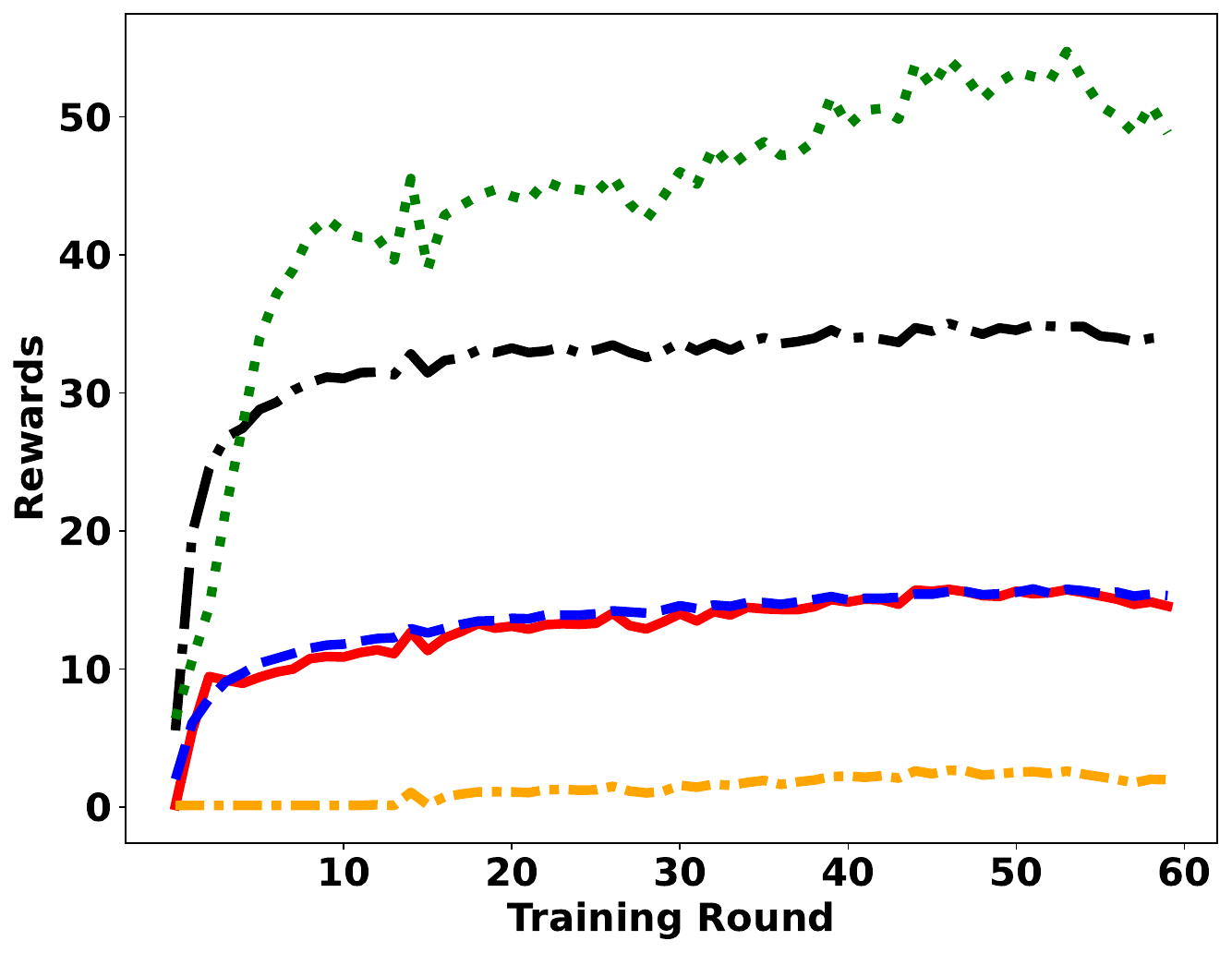}
\caption{\centering Reward distribution per client with different resource allocation}
\end{subfigure}
\hspace{0.06in}
\begin{subfigure}[b]{0.30\textwidth}
\centering
\includegraphics[width=\textwidth]{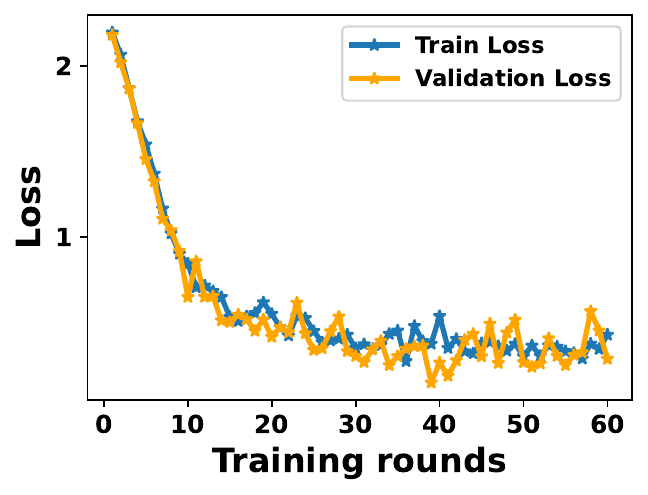}
\caption{\centering Convergence of \texttt{OPUS-VFL} over 60 rounds with 25 clients on MNIST}
\end{subfigure}
\caption{Reward dynamics and model convergence in \texttt{OPUS-VFL}: (a) Reward distribution over training rounds when all five organizations (i.e., clients) allocate the same fraction $\mathcal{C}_i$ of their available resources. (b) Reward distribution when clients allocate different fractions of their resources illustrates our incentive mechanism's fairness. For example, a moderately contributing client with higher resource allocation (e.g., Organization 4) may receive more reward than a higher contributing client with lower resource use (e.g., Organization 2). (c) Global model convergence is maintained despite the injection of differential privacy noise and dynamic adjustment of $\varepsilon_{it}$ by each client to maximize individual rewards.}

\Description{OPUS-VFL mnist-plots}
\label{fig:mnist-plots}
\end{figure*}

\begin{table}[t]
\centering
\caption{\centering Attack Success Rate (ASR) and Model Accuracy Under Varying Poisoning Budgets Across VFL Methods}
\label{tab:asr_accuracy_backdoor}
\small
\begin{tabular}{lcccc}
\toprule
\textbf{Method} & \textbf{Poisoning Budget (pd)} & \textbf{ASR} & \textbf{Accuracy} \\
\midrule
\multirow{4}{*}{VFL-SGD~\cite{shokri2015privacy}} 
  & 0.1 & \textbf{0.1238} & 0.7351 \\
  & 0.2 & 0.2077 & 0.7334 \\
  & 0.3 & 0.2308 & 0.7051 \\
  & 0.5 & 0.4107 & 0.6863 \\
\midrule
\multirow{4}{*}{VFL-CZOFO~\cite{wang2023unified}} 
  & 0.1 & 0.2561 & 0.6972 \\
  & 0.2 & 0.5583 & 0.6966 \\
  & 0.3 & 0.7024 & 0.6868 \\
  & 0.5 & 0.8604 & 0.6861 \\
\midrule
\multirow{4}{*}{Vanilla-VFL~\cite{vepakomma2018split}} 
  & 0.1 & 0.3960 & 0.7334 \\
  & 0.2 & 0.7600 & 0.7193 \\
  & 0.3 & 0.8740 & 0.7301 \\
  & 0.5 & 0.8891 & 0.6003 \\
\midrule
\multirow{4}{*}{\texttt{OPUS-VFL} \textbf{(Ours)}} 
  & 0.1 &0.1943  & 0.6810 \\
  & 0.2 &\textbf{0.2032}  &  0.6540\\
  & 0.3 &\textbf{0.1987}  & 0.6330 \\
  & 0.5 &\textbf{0.1853}  & 0.6350 \\
\midrule
\multirow{4}{*}{BPF~\cite{tan2023fraim}} 
  & 0.1 & 0.3880 & 0.7310 \\
  & 0.2 & 0.7790 & 0.7140 \\
  & 0.3 & 0.8421 & 0.7273 \\
  & 0.5 & 0.8806 & 0.7284 \\
\midrule
\multirow{4}{*}{O2S~\cite{tan2023fraim}} 
  & 0.1 & 0.3410 & 0.7237 \\
  & 0.2 & 0.7146 & 0.7011 \\
  & 0.3 & 0.8498 & 0.7200 \\
  & 0.5 & 0.8909 & 0.7260 \\
\bottomrule
\end{tabular}
\end{table}

\begin{table}[htbp]
    \centering
    \caption{\centering Label Inference Accuracy Across Client Configurations}
    \small
    \begin{tabular}{P{0.8cm}P{0.8cm}P{0.8cm}P{0.8cm}P{0.8cm}P{0.8cm}P{0.8cm}}
    \toprule
    \textbf{Clients} & \textbf{VFL-SGD} & \textbf{VFL-CZOFO} & \textbf{Vanilla} & \textbf{OPUS-VFL}  \textbf{(Ours)} & \textbf{BPF} & \textbf{O2S} \\
    \midrule
    2 & 0.5576 & 0.4572 & 0.7402 & \textbf{0.3908} & 0.7383 & 0.7452 \\
    3 & 0.5332 & 0.4388 & 0.7207 & \textbf{0.3927} & 0.7445 & 0.7286 \\
    4 & 0.5290 & 0.4285 & 0.7077 & \textbf{0.3984} & 0.7146 & 0.7175 \\
    5 & 0.4922 & 0.4136 & 0.6913 & \textbf{0.3961} & 0.6923 & 0.6872 \\
    6 & 0.4835 & 0.3991 & 0.6351 & \textbf{0.3922 }& 0.6246 & 0.6361 \\
    \bottomrule
    \end{tabular}
    \label{tab:label-inference-accuracy}
\end{table}

\begin{table}[htbp]
    \centering
    \caption{\centering Feature Inference MSE Across Client Configurations}
    \small 
    \begin{tabular}{P{0.8cm}P{0.8cm}P{0.8cm}P{0.8cm}P{0.8cm}P{0.8cm}P{0.8cm}}
    \toprule
    \textbf{Clients} & \textbf{VFL-SGD} & \textbf{VFL-CZOFO} & \textbf{Vanilla} & \textbf{OPUS-VFL}  \textbf{(Ours)} & \textbf{BPF} & \textbf{O2S} \\
    \midrule
    2 & 6.74 & 5.79 & 4.30 & \textbf{6.85} & 5.01 & 4.90 \\
    3 & 6.70 & 5.60 & 4.49 & \textbf{16.84} & 4.75 & 4.68 \\
    4 & 6.50 & 5.55 & 5.58 & \textbf{47.57} & 6.58 & 5.50 \\
    5 & 6.25 & 5.40 & 6.57 & \textbf{78.80} & 6.48 & 6.43 \\
    6 & 6.10 & 5.25 & 7.28 & \textbf{79.64} & 7.23 & 7.15 \\
    \bottomrule
    \end{tabular}
    \label{tab:feature-inference-mse}
\end{table}

\subsection{Robustness of \texttt{OPUS-VFL} Under Adversarial Attacks}
\label{sec:p2vfl-against-attacks}

\subsubsection{\bf Robustness Against Data Poisoning Attacks (Backdoor Attacks)}
\textbf{Table ~\ref{tab:asr_accuracy_backdoor}} presents the experimental results evaluating the robustness of the VFL schemes against backdoor attacks under varying poisoning budgets, in terms of ASR and prediction accuracy.  \texttt{OPUS-VFL} consistently outperforms all baselines in terms of robustness, achieving the lowest Attack Success Rate (ASR) across all poisoning budgets. At the highest poisoning budget (pd = 0.5), \texttt{OPUS-VFL} reaches an ASR of just 0.185, compared to 0.889 for Vanilla-VFL, 0.880 for BPF, and 0.860 for VFL-CZOFO. Even at lower budgets (e.g., pd = 0.1), \texttt{OPUS-VFL} maintains a significantly lower ASR (0.194) than Vanilla-VFL (0.396), BPF (0.388), and O2S (0.341), demonstrating consistent resistance to poisoning attacks.

On average, across all attack strengths, \texttt{OPUS-VFL} achieves the lowest mean ASR (0.1954), outperforming BPF (0.7224), VFL-CZOFO (0.5943), and O2S (0.6991). While VFL-SGD shows a slightly higher average ASR (0.2432), it becomes notably less robust at higher poisoning levels, reaching 0.411 at pd = 0.5, highlighting its instability under aggressive attacks.

Although \texttt{OPUS-VFL} yields slightly lower clean accuracy than some baselines, its significantly reduced ASR under attack makes it the more favorable choice in scenarios where robustness is critical. The trade-off in accuracy is minimal compared to the substantial gains in security and model integrity. Moreover, the stability of \texttt{OPUS-VFL}'s ASR across increasing poisoning budgets underscores its effectiveness as a secure and practical solution for real-world VFL deployments.

\subsubsection{\bf Robustness to Label Inference Attacks (LIAs)}  \textbf{Table~\ref{tab:label-inference-accuracy}} presents the robustness of \texttt{OPUS-VFL} against label inference attacks, compared to other VFL methods.  \texttt{OPUS-VFL} consistently achieves lower adversarial success rates in label inference attacks than other baseline methods. For instance, while Vanilla-VFL, BPF, and O2S allow attacker accuracies ranging from 63\% to 74\%, \texttt{OPUS-VFL} holds this rate to approximately 39\%. This 20–30 percentage point reduction indicates that \texttt{OPUS-VFL} more effectively obscures label information, leaving the attacker’s performance near the level of random guessing in a 10-class classification setting.

This improvement can be attributed to \texttt{OPUS-VFL}’s strategy of injecting substantial noise in every training round, which limits direct gradient-based leakage of label-relevant signals and enhances privacy against inference attacks.

    

\subsubsection{\bf Robustness Against Feature Inference Attacks (FIAs)} \textbf{Table~\ref{tab:feature-inference-mse}} presents the robustness of \texttt{OPUS-VFL} against feature inference attacks in comparison with other VFL baselines.  Against feature inference attacks, \texttt{OPUS-VFL} yields substantially higher mean squared error (MSE) values, indicating lower reconstruction quality for adversaries. 
While other baselines generally keep MSE values below 8.0 across all client configurations, OPUS-VFL ranges from 6.85 (for two clients) up to 79.64 (for six clients), leading to a substantial increase in reconstruction error compared to the other methods.

In VFL-SGD, full model updates are exchanged between parties, allowing malicious actors to exploit gradient signals or parameter differences to infer sensitive information. Similarly, although VFL-CZOFO uses a zero-order approach to reduce direct gradient exposure, it still reveals enough parameter-level or partial derivative information for adversaries to approximate inputs or labels.

In contrast, \texttt{OPUS-VFL} selectively masks and localizes updates, sharing only a narrow slice of model parameters or intermediate representations at each iteration. This design inherently limits the attacker’s ability to perform accurate inversion attacks, resulting in lower label inference accuracy (LIA) and higher feature inference error (FIA). Consequently, \texttt{OPUS-VFL} enforces stronger isolation of each participant’s local data, offering significantly better privacy than the more direct update-sharing protocols used in VFL-SGD and VFL-CZOFO.

\section{Conclusions \& Future Work} \label{subsec:conclusion-future-work}

To conclude, we summarize the \textbf{three key contributions} of this work, underscoring the technical merits of \texttt{OPUS-VFL}. First, it introduces a privacy-aware contribution evaluation method that dynamically and efficiently measures each client’s impact in every training round. By leveraging a lightweight leave-one-out (LOO) strategy combined with customizable differential privacy, \texttt{OPUS-VFL} avoids the computational burden of Shapley value or retraining-based methods while ensuring accurate and privacy-preserving assessments.  Second, \texttt{OPUS-VFL} advances incentive design by incorporating resource-sensitive contribution equity. Unlike prior approaches that focus solely on performance gains, our framework also accounts for the resource investment of each client, promoting fairness and enabling participation from data-rich clients with limited computational capacity.  Third, the framework demonstrates strong scalability and efficiency. By avoiding heavy game-theoretic or Shapley-based computations, \texttt{OPUS-VFL} remains practical and robust as the number of clients increases. Its streamlined protocol supports real-world deployment in privacy-critical and heterogeneous federated learning settings, such as healthcare and finance.

The summary of our \textbf{key findings} from the experimental study is as follows. \texttt{OPUS-VFL} significantly outperforms conventional baselines such as VFL-SGD and VFL-CZOFO regarding training efficiency, scalability, and privacy preservation. By selectively sharing model parameters, restricting gradient exposure, and dynamically tuning privacy settings, it achieves strong robustness against label and feature inference attacks while maintaining competitive accuracy. The global model successfully converges even when clients inject differential privacy noise and adjust their privacy budgets over time to optimize individual rewards. Additionally, our incentive mechanism ensures fairness by aligning reward distribution with performance contribution and resource allocation, enabling equitable outcomes even for resource-constrained clients.

\textbf{Future work directions} include incorporating mechanisms to ensure client truthfulness, which can strengthen security by promoting honest reporting. Enhancing model accuracy through dynamic collaborative feature selection and adapting shared features during training may lead to better generalization. Addressing straggler effects via stale-tolerant training can improve robustness in real-world deployments. Additionally, integrating fairness constraints such as demographic parity or equal opportunity can support equitable outcomes. These directions will further refine \texttt{OPUS-VFL} and extend its impact in collaborative learning settings.

\newpage

\bibliographystyle{plainnat}  
\bibliography{main}

@article{yang2019federated,
  title={Federated machine learning: Concept and applications},
  author={Yang, Qiang and Liu, Yang and Chen, Tianjian and Tong, Yongxin},
  journal={ACM Transactions on Intelligent Systems and Technology (TIST)},
  volume={10},
  number={2},
  pages={1--19},
  year={2019},
  publisher={ACM New York, NY, USA}
}

@article{zhang2021survey,
  title={A survey on federated learning},
  author={Zhang, Chen and Xie, Yu and Bai, Hang and Yu, Bin and Li, Weihong and Gao, Yuan},
  journal={Knowledge-Based Systems},
  volume={216},
  pages={106775},
  year={2021},
  publisher={Elsevier}
}

@inproceedings{tan2023fraim,
  title={FRAIM: A Feature Importance-Aware Incentive Mechanism for Vertical Federated Learning},
  author={Tan, Lei and Yang, Yunchao and Hu, Miao and Zhou, Yipeng and Wu, Di},
  booktitle={International Conference on Algorithms and Architectures for Parallel Processing},
  pages={132--150},
  year={2023},
  organization={Springer}
}

@article{wang2024unified,
  title={A unified solution for privacy and communication efficiency in vertical federated learning},
  author={Wang, Ganyu and Gu, Bin and Zhang, Qingsong and Li, Xiang and Wang, Boyu and Ling, Charles X},
  journal={Advances in Neural Information Processing Systems},
  volume={36},
  year={2024}
}

@inproceedings{li2023fedsdg,
  title={FedSDG-FS: Efficient and secure feature selection for vertical federated learning},
  author={Li, Anran and Peng, Hongyi and Zhang, Lan and Huang, Jiahui and Guo, Qing and Yu, Han and Liu, Yang},
  booktitle={IEEE INFOCOM 2023-IEEE Conference on Computer Communications},
  pages={1--10},
  year={2023},
  organization={IEEE}
}

@article{yang2023practical,
  title={Practical feature inference attack in vertical federated learning during prediction in artificial Internet of Things},
  author={Yang, Ruikang and Ma, Jianfeng and Zhang, Junying and Kumari, Saru and Kumar, Sachin and Rodrigues, Joel JPC},
  journal={IEEE Internet of Things Journal},
  volume={11},
  number={1},
  pages={5--16},
  year={2023},
  publisher={IEEE}
}

@article{lu2022truthful,
  title={Truthful incentive mechanism design via internalizing externalities and {LP} relaxation for vertical federated learning},
  author={Lu, Jianfeng and Pan, Bangqi and Seid, Abegaz Mohammed and Li, Bing and Hu, Gangqiang and Wan, Shaohua},
  journal={IEEE Transactions on Computational Social Systems},
  year={2022},
  publisher={IEEE}
}

@article{liu2024vertical,
  title={Vertical federated learning: Concepts, advances, and challenges},
  author={Liu, Yang and Kang, Yan and Zou, Tianyuan and Pu, Yanhong and He, Yuanqin and Ye, Xiaozhou and Ouyang, Ye and Zhang, Ya-Qin and Yang, Qiang},
  journal={IEEE Transactions on Knowledge and Data Engineering},
  year={2024},
  publisher={IEEE}
}

@inproceedings{zhang-horizontal-2021,
  author = {Zhang, Jingwen and Wu, Yuezhou and Pan, Rong},
  title = {Incentive mechanism for horizontal federated learning based on reputation and reverse auction},
  year = {2021},
  month = {Apr.},
  booktitle = {Proceedings of the Web Conference 2021}
}

@INPROCEEDINGS{deng-quality-aware-2021,
  author={Deng, Yongheng and Lyu, Feng and Ren, Ju and Chen, Yi-Chao and Yang, Peng and Zhou, Yuezhi and Zhang, Yaoxue},
  booktitle={IEEE IEEE Conference on Computer Communications (INFOCOM)}, 
  title={{FAIR}: Quality-aware federated learning with precise user incentive and model aggregation}, 
  year={2021},
  publisher={IEEE}
  }

@INPROCEEDINGS{Han22-tiff,  author={Han, Jingoo and Khan, Ahmad Faraz and Zawad, Syed and Anwar, Ali and Angel, Nathalie Baracaldo and Zhou, Yi and Yan, Feng and Butt, Ali R.},  booktitle={2022 IEEE 15th International Conference on Cloud Computing (CLOUD)},   title={{TIFF}: Tokenized Incentive for Federated Learning},   year={2022},  volume={},  number={},  pages={407-416},  doi={10.1109/CLOUD55607.2022.00064}}

@INPROCEEDINGS{Toyoda19,  
author={Kentaroh Toyoda and Allan N. Zhang},
booktitle={2019 IEEE International Conference on Big Data (Big Data)},   
title={Mechanism Design for An Incentive-aware Blockchain-enabled Federated Learning Platform},   
year={2019},  
pages={395--403},  
doi={10.1109/BigData47090.2019.9006344}
}

@INPROCEEDINGS{tang-incentive-2021,  author={Tang, Ming and Wong, Vincent W.S.},  booktitle={IEEE INFOCOM 2021 - IEEE Conference on Computer Communications},   title={An Incentive Mechanism for Cross-Silo Federated Learning: A Public Goods Perspective},   year={2021},  volume={},  number={},  pages={1-10},  doi={10.1109/INFOCOM42981.2021.9488705}}

@inproceedings{ng2021multi,
  title={A multi-player game for studying federated learning incentive schemes},
  author={Ng, Kang Loon and Chen, Zichen and Liu, Zelei and Yu, Han and Liu, Yang and Yang, Qiang},
  booktitle={Proceedings of the Twenty-Ninth International Conference on International Joint Conferences on Artificial Intelligence (IJCAI)},
  pages={5279--5281},
  year={2021}
}

@inproceedings{khan2024using,
  title={Using the nucleolus for incentive allocation in vertical federated learning},
  author={Khan, Afsana and ten Thij, Marijn and Thuijsman, Frank and Wilbik, Anna},
  booktitle={2024 2nd International Conference on Federated Learning Technologies and Applications (FLTA)},
  
  year={2024}
}

@inproceedings{yang2023incentive,
  title={Incentive Mechanism Design for Vertical Federated Learning},
  author={Yang, Ni and Cheung, Man Hon},
  booktitle={ICC 2023-IEEE International Conference on Communications},
  year={2023}
}

@inproceedings{sun2024hifi,
  title={{HiFi-Gas}: Hierarchical federated learning incentive mechanism enhanced gas usage estimation},
  author={Sun, Hao and Tang, Xiaoli and Yang, Chengyi and Yu, Zhenpeng and Wang, Xiuli and Ding, Qijie and Li, Zengxiang and Yu, Han},
  booktitle={Proceedings of the AAAI Conference on Artificial Intelligence},
  year={2024}
}

@article{cui2024survey,
  title={A Survey on Contribution Evaluation in Vertical Federated Learning},
  author={Cui, Yue and Huang, Chung-ju and Zhang, Yuzhu and Wang, Leye and Fan, Lixin and Zhou, Xiaofang and Yang, Qiang},
  journal={arXiv preprint arXiv:2405.02364},
  year={2024}
}

@article{le2021incentive,
  title={An incentive mechanism for federated learning in wireless cellular networks: An auction approach},
  author={Le, Tra Huong Thi and Tran, Nguyen H and Tun, Yan Kyaw and Nguyen, Minh NH and Pandey, Shashi Raj and Han, Zhu and Hong, Choong Seon},
  journal={IEEE Transactions on Wireless Communications},
  volume={20},
  number={8},
  pages={4874--4887},
  year={2021},
  publisher={IEEE}
}

@inproceedings{zhang2021incentive,
  title={Incentive mechanism for horizontal federated learning based on reputation and reverse auction},
  author={Zhang, Jingwen and Wu, Yuezhou and Pan, Rong},
  booktitle={Proceedings of the Web Conference 2021},
  pages={947--956},
  year={2021}
}

@article{deng2022improving,
  title={Improving federated learning with quality-aware user incentive and auto-weighted model aggregation},
  author={Deng, Yongheng and Lyu, Feng and Ren, Ju and Chen, Yi-Chao and Yang, Peng and Zhou, Yuezhi and Zhang, Yaoxue},
  journal={IEEE Transactions on Parallel and Distributed Systems},
  volume={33},
  number={12},
  pages={4515--4529},
  year={2022},
  publisher={IEEE}
}

@article{ng2021hierarchical,
  title={A hierarchical incentive design toward motivating participation in coded federated learning},
  author={Ng, Jer Shyuan and Lim, Wei Yang Bryan and Xiong, Zehui and Cao, Xianbin and Niyato, Dusit and Leung, Cyril and Kim, Dong In},
  journal={IEEE Journal on Selected Areas in Communications},
  volume={40},
  number={1},
  pages={359--375},
  year={2021},
  publisher={IEEE}
}

@inproceedings{yan2021fedcm,
  title={Fedcm: A real-time contribution measurement method for participants in federated learning},
  author={Yan, Bingjie and Liu, Boyi and Wang, Lujia and Zhou, Yize and Liang, Zhixuan and Liu, Ming and Xu, Cheng-Zhong},
  booktitle={2021 International joint conference on neural networks (IJCNN)},
  pages={1--8},
  year={2021},
  organization={IEEE}
}

@article{xu2024elxgb,
  title={ELXGB: An efficient and privacy-preserving XGBoost for vertical federated learning},
  author={Xu, Wei and Zhu, Hui and Zheng, Yandong and Wang, Fengwei and Zhao, Jiaqi and Liu, Zhe and Li, Hui},
  journal={IEEE Transactions on Services Computing},
  year={2024},
  publisher={IEEE}
}

@article{fan2024securevfl,
  title={SecureVFL: privacy-preserving multi-party vertical federated learning based on blockchain and RSS},
  author={Fan, Mochan and Zhang, Zhipeng and Li, Zonghang and Sun, Gang and Yu, Hongfang and Kang, Jiawen and Guizani, Mohsen},
  journal={Digital Communications and Networks},
  year={2024},
  publisher={Elsevier}
}

@inproceedings{xu2021fedv,
  title={Fedv: Privacy-preserving federated learning over vertically partitioned data},
  author={Xu, Runhua and Baracaldo, Nathalie and Zhou, Yi and Anwar, Ali and Joshi, James and Ludwig, Heiko},
  booktitle={Proceedings of the 14th ACM workshop on artificial intelligence and security},
  pages={181--192},
  year={2021}
}

@inproceedings{li2023fedvs,
  title={FedVS: Straggler-resilient and privacy-preserving vertical federated learning for split models},
  author={Li, Songze and Yao, Duanyi and Liu, Jin},
  booktitle={International Conference on Machine Learning},
  pages={20296--20311},
  year={2023},
  organization={PMLR}
}

@article{zhu2021pivodl,
  title={{PIVODL}: Privacy-preserving vertical federated learning over distributed labels},
  author={Zhu, Hangyu and Wang, Rui and Jin, Yaochu and Liang, Kaitai},
  journal={IEEE Transactions on Artificial Intelligence},
  volume={4},
  number={5},
  pages={988--1001},
  year={2021},
  publisher={IEEE}
}

@article{li2023efficient,
  title={Efficient and privacy-preserving feature importance-based vertical federated learning},
  author={Li, Anran and Huang, Jiahui and Jia, Ju and Peng, Hongyi and Zhang, Lan and Tuan, Luu Anh and Yu, Han and Li, Xiang-Yang},
  journal={IEEE Transactions on Mobile Computing},
  volume={23},
  number={6},
  pages={7238--7255},
  year={2023},
  publisher={IEEE}
}

@article{wu2023falcon,
  title={Falcon: A privacy-preserving and interpretable vertical federated learning system},
  author={Wu, Yuncheng and Xing, Naili and Chen, Gang and Dinh, Tien Tuan Anh and Luo, Zhaojing and Ooi, Beng Chin and Xiao, Xiaokui and Zhang, Meihui},
  journal={Proceedings of the VLDB Endowment},
  volume={16},
  number={10},
  pages={2471--2484},
  year={2023},
  publisher={VLDB Endowment}
}

@inproceedings{shokri2015privacy,
  title={Privacy-preserving deep learning},
  author={Shokri, Reza and Shmatikov, Vitaly},
  booktitle={Proceedings of the 22nd ACM SIGSAC conference on computer and communications security},
  pages={1310--1321},
  year={2015}
}

@article{wang2023unified,
  title={A unified solution for privacy and communication efficiency in vertical federated learning},
  author={Wang, Ganyu and Gu, Bin and Zhang, Qingsong and Li, Xiang and Wang, Boyu and Ling, Charles X},
  journal={Advances in Neural Information Processing Systems},
  volume={36},
  pages={13480--13491},
  year={2023}
}

@inproceedings{xie2024improving,
  title={Improving privacy-preserving vertical federated learning by efficient communication with admm},
  author={Xie, Chulin and Chen, Pin-Yu and Li, Qinbin and Nourian, Arash and Zhang, Ce and Li, Bo},
  booktitle={2024 IEEE Conference on Secure and Trustworthy Machine Learning (SaTML)},
  pages={443--471},
  year={2024},
  organization={IEEE}
}

@article{vepakomma2018split,
  title={Split learning for health: Distributed deep learning without sharing raw patient data},
  author={Vepakomma, Praneeth and Gupta, Otkrist and Swedish, Tristan and Raskar, Ramesh},
  journal={arXiv preprint arXiv:1812.00564},
  year={2018}
}

@article{zhang2024privacy,
  title={Privacy-preserving Data Selection for Horizontal and Vertical Federated Learning},
  author={Zhang, Lan and Li, Anran and Peng, Hongyi and Han, Feng and Huang, Fan and Li, Xiang-Yang},
  journal={IEEE Transactions on Parallel and Distributed Systems},
  year={2024},
  publisher={IEEE}
}

@article{zheng2023privet,
  title={Privet: A privacy-preserving vertical federated learning service for gradient boosted decision tables},
  author={Zheng, Yifeng and Xu, Shuangqing and Wang, Songlei and Gao, Yansong and Hua, Zhongyun},
  journal={IEEE Transactions on Services Computing},
  volume={16},
  number={5},
  pages={3604--3620},
  year={2023},
  publisher={IEEE}
}

@article{wang2025pravfed,
  title={Pravfed: Practical heterogeneous vertical federated learning via representation learning},
  author={Wang, Shuo and Gai, Keke and Yu, Jing and Zhang, Zijian and Zhu, Liehuang},
  journal={IEEE Transactions on Information Forensics and Security},
  year={2025},
  publisher={IEEE}
}

@article{yang2024openvfl,
  title={OpenVFL: A Vertical Federated Learning Framework with Stronger Privacy-Preserving},
  author={Yang, Yunbo and Chen, Xiang and Pan, Yuhao and Shen, Jiachen and Cao, Zhenfu and Dong, Xiaolei and Li, Xiaoguo and Sun, Jianfei and Yang, Guomin and Deng, Robert},
  journal={IEEE Transactions on Information Forensics and Security},
  year={2024},
  publisher={IEEE}
}

@article{wen2023survey,
  title={A survey on federated learning: challenges and applications},
  author={Wen, Jie and Zhang, Zhixia and Lan, Yang and Cui, Zhihua and Cai, Jianghui and Zhang, Wensheng},
  journal={International Journal of Machine Learning and Cybernetics},
  volume={14},
  number={2},
  pages={513--535},
  year={2023},
  publisher={Springer}
}

@inproceedings{abadi2016deep,
  title={Deep learning with differential privacy},
  author={Abadi, Martin and Chu, Andy and Goodfellow, Ian and McMahan, H Brendan and Mironov, Ilya and Talwar, Kunal and Zhang, Li},
  booktitle={Proceedings of the 2016 ACM SIGSAC conference on computer and communications security},
  pages={308--318},
  year={2016}
}

@inproceedings{fu2022label,
  title={Label inference attacks against vertical federated learning},
  author={Fu, Chong and Zhang, Xuhong and Ji, Shouling and Chen, Jinyin and Wu, Jingzheng and Guo, Shanqing and Zhou, Jun and Liu, Alex X and Wang, Ting},
  booktitle={31st USENIX security symposium (USENIX Security 22)},
  pages={1397--1414},
  year={2022}
}

@inproceedings{naseri2024badvfl,
  title={Badvfl: Backdoor attacks in vertical federated learning},
  author={Naseri, Mohammad and Han, Yufei and De Cristofaro, Emiliano},
  booktitle={2024 IEEE Symposium on Security and Privacy (SP)},
  pages={2013--2028},
  year={2024},
  organization={IEEE}
}

\newpage
\appendices
\section{Additional Experimental Results}


\begin{table}[h]
\centering
\caption{\centering Performance Metrics for Different $\varepsilon$ Values}
\label{tab:eps-ablation-performance}
\begin{tabular}{P{1cm}P{2cm}P{2cm}}
\toprule
$\varepsilon$ & Train Acc. & Test Acc. \\
\midrule
0.1 & 0.8492 & 0.8131 \\
0.3 & 0.8970 & 0.9230 \\
0.5 & 0.9191 & 0.9120 \\
0.7 & 0.9007 & 0.9340 \\
0.9 & 0.9227 & 0.9010 \\
1.0 & 0.9154 & 0.9450 \\
\bottomrule
\end{tabular}
\end{table}

\textbf{Table~\ref {tab:eps-ablation-performance}} presents a comparative analysis of \texttt{OPUS-VFL} model performance across varying privacy budgets, parameterized by $\varepsilon$.  Specifically, the table reports both training and testing accuracies for each tested value of $\varepsilon$.  Notably, intermediate $\varepsilon$=0.7 achieves a favorable balance, yielding higher test accuracies (0.9230 and 0.9340, respectively) than more conservative settings. The highest test accuracy (0.9450) is observed at $\varepsilon=1.0$, despite a marginal reduction in training performance compared to some lower $\varepsilon$ settings, suggesting that moderate relaxation of privacy constraints might enhance model generalization. 

\begin{table}[ht]
\centering
\caption{\centering Experimental Results for Varying $\tau_i$, Dropout Clients, and Warmup Epochs}
\label{tab:results-tau-varying}
\begin{tabular}{P{1cm}P{1cm}P{1cm}P{1cm}P{1cm}}
\toprule
$\tau_i$ & Dropout Clients & Warmup Epochs & Total Epochs & Test Acc. \\
\midrule
10   & 15  & 60  & 150 & 0.9067 \\
100  & 0   & 60  & 150 & 0.9123 \\
120  & 0   & 60  & 150 & 0.9411 \\
500  & 0   & 60  & 150 & 0.9558 \\
50   & 6   & 40  & 150 & 0.9448 \\
70   & 7   & 40  & 150 & 0.9411 \\
90   & 1   & 40  & 150 & 0.9154 \\
110  & 0   & 40  & 150 & 0.9227 \\
\bottomrule
\end{tabular}
\end{table}

\textbf{Table~\ref{tab:results-tau-varying}} summarizes the test accuracies achieved under varying server budget allocations for clients. To simulate client drop-out scenarios, we include a warm-up phase during which the model stabilizes before enabling client dropouts. When client dropout is triggered earlier, smaller server budgets lead to more clients leaving the federation. In such cases, a higher server budget is required to maintain effective training. Conversely, under lower server budgets, extending the warm-up phase allows clients more time to accumulate utility, reducing dropout and improving overall model performance.

\begin{figure}[h]
    \centering
    \includegraphics[width=0.35\textwidth]{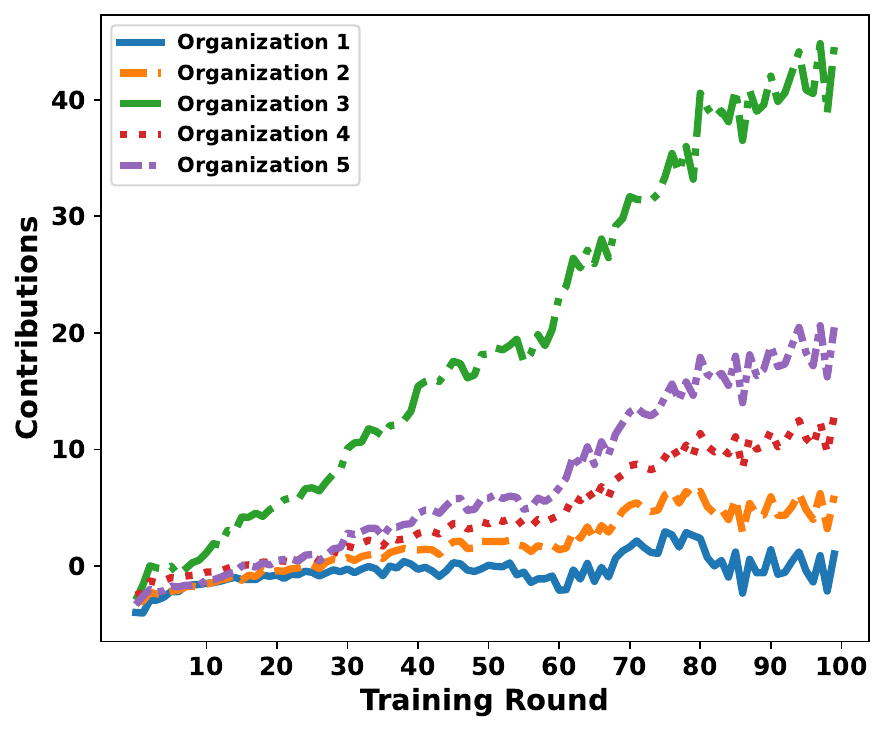}
\caption{Contributions on the CIFAR-10 dataset over 100 training rounds for 5 clients under a backdoor poisoning budget 0.5.}
    \label{fig:cifar10-contributions-backdoor}
\end{figure}

\textbf{Figure~\ref{fig:cifar10-contributions-backdoor}} shows client contributions on the CIFAR10 dataset under a backdoor attack with a poisoning budget of 0.5. Notably, the contribution patterns remain largely consistent with those in the clean setting (see Figure~\ref{fig:cifar10-plots}), exhibiting minimal deviation even under a high poisoning budget.

\begin{figure}[h]
    \centering
    \includegraphics[width=0.35\textwidth]{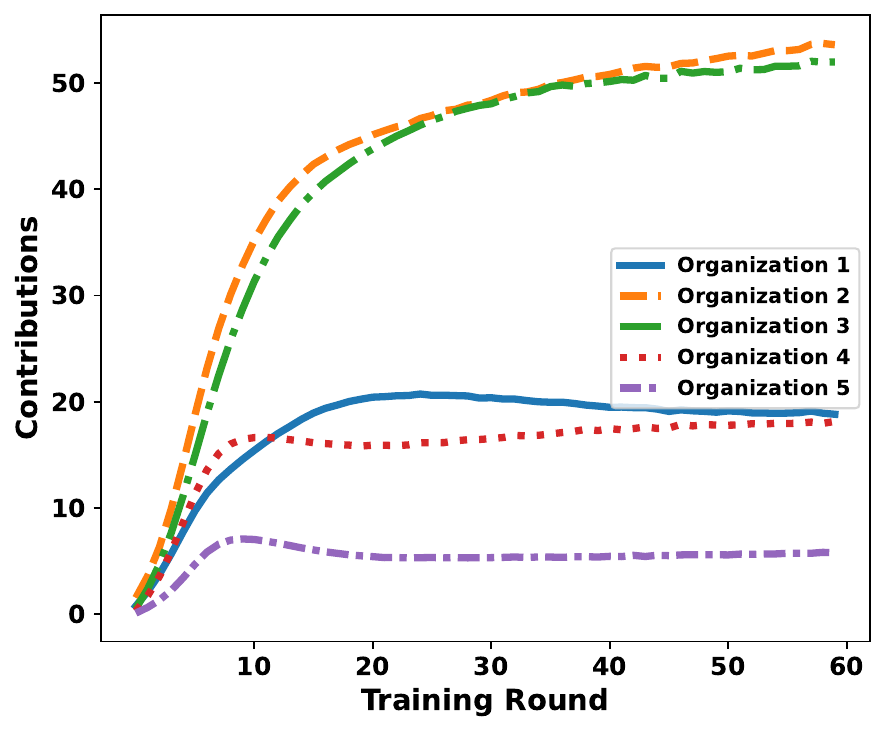}
\caption{Contributions on the MINIST dataset when the step size for changing $\varepsilon$ is 0 and $\varepsilon$ is initialized to $10^6$, indicating negligible noise.
}
    \label{fig:mnist-without-privacy}
\end{figure}

\textbf{Figure~\ref{fig:mnist-without-privacy}} depicts the evolution of client contributions over 60 training rounds on the MNIST dataset without noise addition. The plots closely resemble those with noise (see Figure~\ref{fig:mnist-plots}), suggesting that the loss difference with and without noise remains minimal when $\varepsilon$ is between 0.5 and 1.0.

\begin{table}[ht]
\centering
\caption{\centering SNR Values and Model Performance per Organization (MNIST with 5 Clients)}
\label{tab:snr-table}
\begin{tabular}{lr}
\toprule
\textbf{Organization} & \textbf{SNR (dB)} \\
\midrule
Org 1 & -2.11 \\
Org 2 & -0.78 \\
Org 3 & -0.93 \\
Org 4 & -1.41 \\
Org 5 & -3.26 \\
\midrule
\textbf{Train Accuracy} & 0.9057 \\
\textbf{Test Accuracy} & 0.91042 \\
\bottomrule
\end{tabular}
\end{table}

\textbf{Table~\ref{tab:snr-table}} presents the SNR values for each of the five client organizations in our \texttt{OPUS-VFL} framework. Negative SNR values, ranging from -0.78 dB to -3.26 dB, indicate that the added noise exceeds the magnitude of the underlying signal. This level of noise serves as an effective privacy measure, demonstrating \texttt{OPUS-VFL}'s ability to obscure sensitive features.

\begin{figure*}  
\centering
\begin{subfigure}[b]{0.3\textwidth}
\centering
\includegraphics[width=\textwidth]{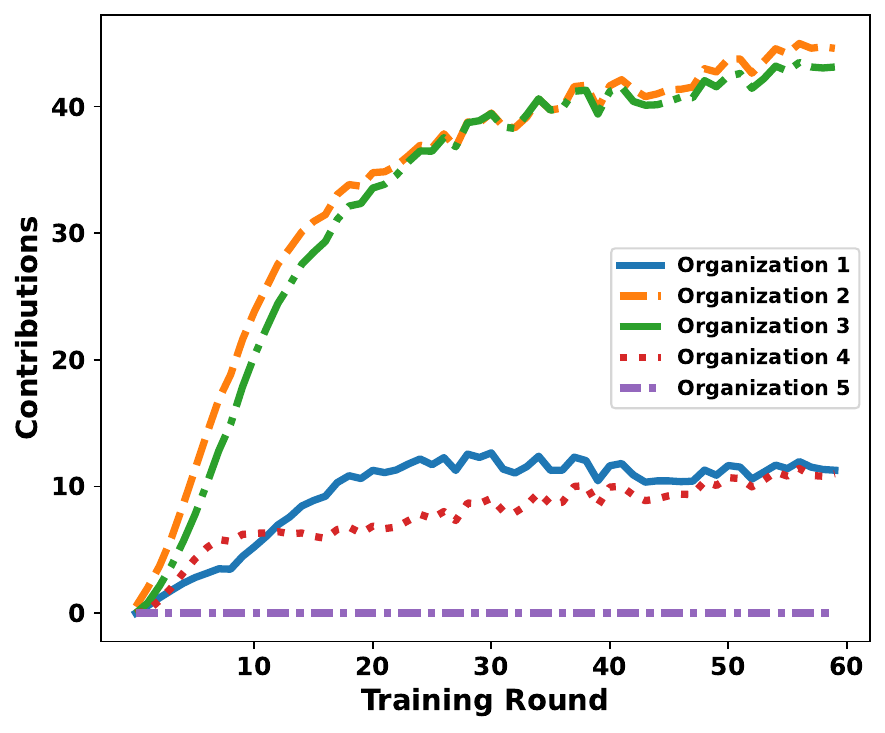}
\caption{Contributions}
\end{subfigure}
\begin{subfigure}[b]{0.3\textwidth}
\centering
\includegraphics[width=\textwidth]{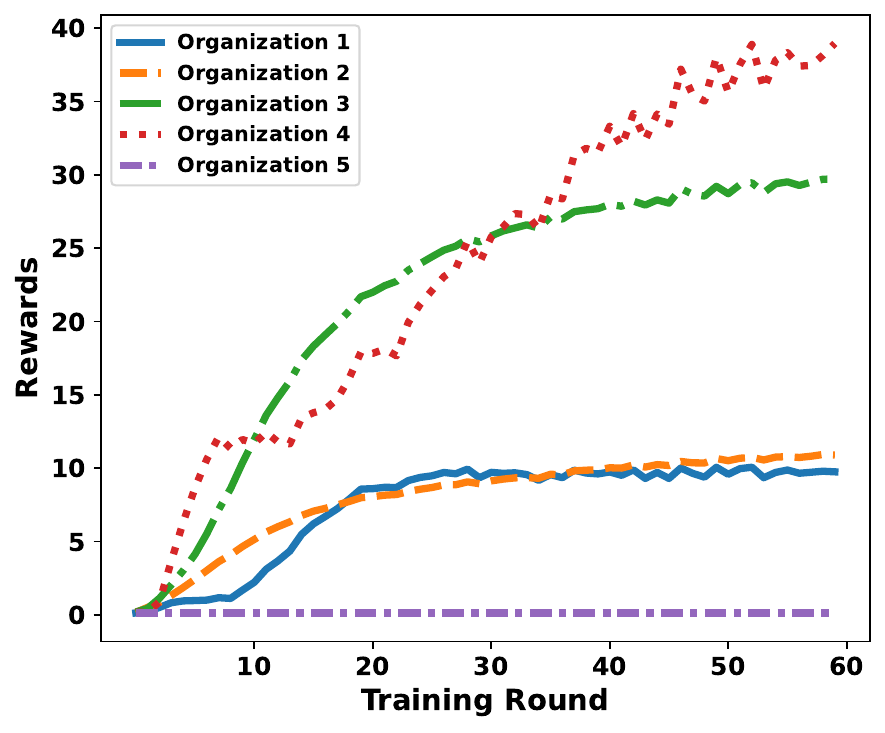}
\caption{Rewards}
\end{subfigure}
\begin{subfigure}[b]{0.3\textwidth}
\centering
\includegraphics[width=\textwidth]{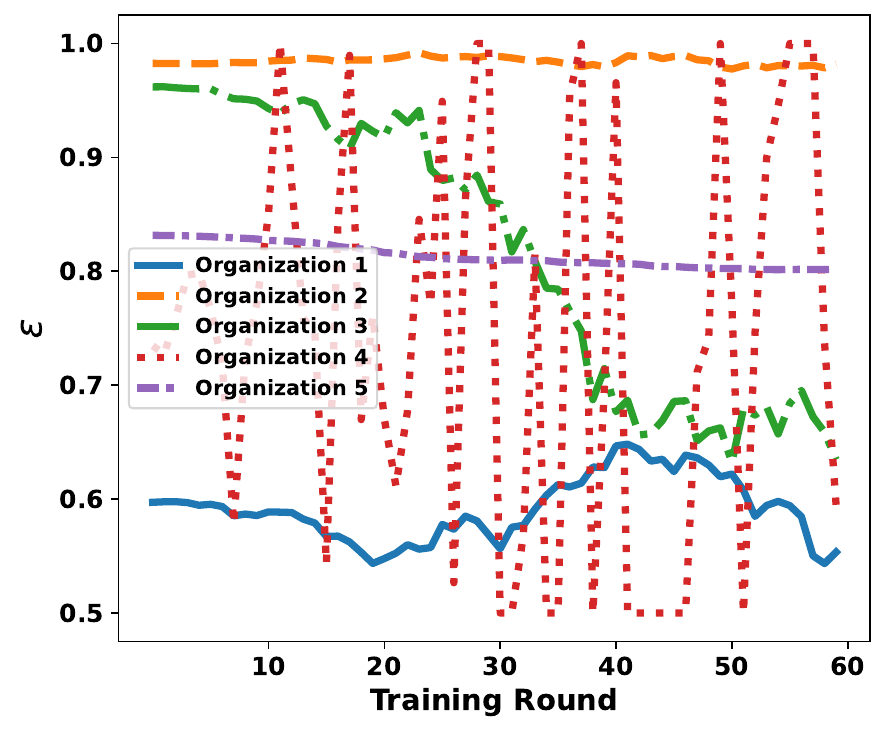}
\caption{Adopted $\varepsilon$ (DP parameter)}
\end{subfigure}
\caption{Contribution-Reward Dynamics and Privacy Adaptation in \texttt{OPUS-VFL}: Contributions, rewards, and differential privacy parameter $\varepsilon$ values generated on the MNIST dataset over 60 training rounds, where $\varepsilon$ dynamically changes across epochs to accommodate high-contributing clients. A configurable step size parameter controls the magnitude of $\varepsilon$ adjustment.}

\Description{OPUS-VFL training plots}
\label{fig:mnist-contributions-rewards-epsilon-plots}
\end{figure*}

\textbf{Figure ~\ref{fig:mnist-contributions-rewards-epsilon-plots}} show the evolution of contributions across all training arounds for MNIST dataset using 5 clients. Each clients adjust their privacy parameter to maximize their rewards and at the same to contribute meaningfully to the global model prediction accuracy.

\end{document}